\documentclass[preprint,12pt,authoryear]{elsarticle}

\usepackage[utf8]{inputenc}
\usepackage[T1]{fontenc}
\usepackage{lmodern}

\usepackage{
	afterpage,
	algorithm,
	algpseudocode,
	amsmath,
	amssymb,
	booktabs,
	lineno,
	longtable,
	graphicx,
	multirow,
	pdflscape,
	subcaption,
	url
}

\mathchardef\mhyphen="2D 

\journal{Computer Speech and Language}

\begin{document}


\begin{frontmatter}



This is a submission of accepted manuscript published by Elsevier in Computer Speech \&
Language, available online https://doi.org/10.1016/j.csl.2017.10.002

\title{Searching for Discriminative Words in Multidimensional Continuous Feature Space}


\author{Marius Sajgalik}
\author{Michal Barla}
\author{Maria Bielikova}

\address{Slovak University of Technology in Bratislava\\
Faculty of Informatics and Information Technologies\\
Ilkovičova 2, 842 16 Bratislava, Slovakia}

\begin{abstract}
Word feature vectors have been proven to improve many natural language processing tasks. With recent advances in unsupervised learning of these feature vectors, it became possible to train it with much more data, which also resulted in better quality of learned features. Since it learns joint probability of latent features of words, it has the advantage that we can train it without any prior knowledge about the goal task we want to solve. We aim to evaluate the universal applicability property of feature vectors, which has been already proven to hold for many standard NLP tasks like part-of-speech tagging or syntactic parsing. In our case, we want to understand the topical focus of text documents and design an efficient representation suitable for discriminating different topics. The discriminativeness can be evaluated adequately on text categorisation task. We propose a novel method to extract discriminative keywords from documents. We utilise word feature vectors to understand the relations between words better and also understand the latent topics which are discussed in the text and not mentioned directly but inferred logically. We also present a simple way to calculate document feature vectors out of extracted discriminative words. We evaluate our method on the four most popular datasets for text categorisation. We show how different discriminative metrics influence the overall results. We demonstrate the effectiveness of our approach by achieving state-of-the-art results on text categorisation task using just a small number of extracted keywords. We prove that word feature vectors can substantially improve the topical inference of documents' meaning. We conclude that distributed representation of words can be used to build higher levels of abstraction as we demonstrate and build feature vectors of documents. Our method can help in any multi-domain environment to automatically extract discriminative keywords. It can be used to organise and search documents more efficiently.
\end{abstract}

\begin{keyword}
	text categorisation \sep
	distributed representation \sep
	feature vectors \sep
	word vectors \sep
	NLP



\end{keyword}

\end{frontmatter}


\section{Introduction}
\label{sec-introduction}

A prerequisite for automated processing of textual information is understanding of individual words and relations between them.
We cannot shift to the understanding of complex sentences if we are still failing to understand simple words.

To determine the meaning of a word we usually try to find a mapping of this word to some concept within some conceptual model we are using (e.g., to determine a synset in Wordnet \citep{Miller:1995:WLD:219717.219748}). Such a mapping, if we can find it, gives us relations of this word to other concepts.
This is the reason why many researchers have been trying to enhance the automated understanding of a raw text by creating hand-crafted dictionaries, thesauri, word taxonomies and ontologies \citep{Hennig:2008:OAT:1487207.1487345}.
Ontologies, in particular, were considered as a solution with high potential, since they utilise a formal apparatus to describe properties of words or concepts and relations between them with great detail and precision.

However, when dealing with text processing tasks, finding the mapping of a word to some ontology concept is not straightforward and easy, since it requires us to solve a word sense disambiguation task, whose low-performance \citep{Navigli:2012:BAC:2397213.2397579}, particularly at lower levels of abstraction, limits the usability of ontologies.

Even if we were able to find the mapping of a word within an ontology, the relations between words that we would get are often unsatisfactory.
Human-crafted ontologies contain only a limited, not necessarily optimal set of relations, which their creators considered as important.
Most of them are described just qualitatively, which is not suitable for multiple common NLP tasks, such as measuring word similarity \citep{Pedersen:2010:ICM:1857999.1858046}.
For instance, the approach in \cite{Barla:2009:DTK:1692019.1692050} reveals only a relation type, but it cannot determine the relation quantitatively.
However, we are often interested just in retrieving the most similar words to enrich our understanding of text meaning.
In such cases, we do not care if the most similar words are synonyms or hypernyms.
From the other point of view, when we are using an ontology, and we want to broaden our understanding of a word by retrieving the most similar words, we have to deal with the qualitative information about word relations since most ontology relations are not quantified.
We have no universally best means of retrieving exactly $K$ most similar ontology concepts.
That is the reason why all existing ontology-based methods for measuring (semantic) similarity are just heuristics \citep{Resnik:1995:UIC:1625855.1625914,Jiang:1997:semanticsimilarity,Lin:1998:IDS:645527.657297,Pedersen:2010:ICM:1857999.1858046} and thus limited to achieving only rather imprecise results.

Probably the biggest deficiency of all ontologies is the scalability.
Using ontologies, we are mostly limited to words and some phrases.
However, it is intractable to scale up and represent all sentences, paragraphs, documents or whole domains and relations among them.

With the recent boom of deep learning, new unsupervised methods have been invented which can learn the meaning of words from a raw text, without any annotations \citep{Bengio:2003:NPL:944919.944966}.
These methods map words into multi-dimensional vector space.
They encode every word by a latent feature vector which captures its meaning.
As a consequence, we do not need to disambiguate the exact sense of a word anymore.
The basis for such representation is the distributional hypothesis \citep{Harris:1954:word}, which states that features of a word can be learned based on its context (i.e. surrounding words in a text). We can compensate the disadvantage of having latent nature of features which we are unable to interpret directly by benefits of having a vector space representation of words.
We can employ standard vector operations to solve many interesting tasks, e.g., the addition of vectors to compose meaning of a phrase or cosine distance of vectors as a similarity measure of pairs of words.
Finding the most similar words of a word boils down to finding the closest neighbours of its vector.
We can also find vectors that encode a relationship between a pair of words, e.g., a vector transforming singular form into plural one, etc.
\cite{Mikolov:2013:linguistic} showed that with such a distributed representation of words based on vectors of latent features, encode many semantic, but interestingly also syntactic relations.

Word feature vectors have been already proven to improve results of many NLP tasks \citep{Collobert:2008:UAN:1390156.1390177,Turian:2010:word}.
They are assumed to possess a valuable property that regardless of the task we want to use them for, we can learn them in advance without any prior knowledge of the task itself.
For instance, recent advances in image caption generation use pre-trained word feature vectors to generate phrases describing images \citep{Karpathy:2014:deep}.

In this paper, we propose a novel method to extract relevant keywords of a document.
We utilise word vectors of latent features to build a semantic representation of documents embedded in the same feature space.
Our goal is to obtain a highly discriminative representation of each document, which could be advantageous in some applications.
Our assumption while doing so is that we have some existing categories within our working domain and we care only about the descriptive words that discriminate the categories or topics.
We are not interested in generic words that define the domain as a whole since we know we are working in that particular domain, and therefore it does not have any real information value for us.

We define a difference between a category and a domain for the scope of this paper, as we use these terms frequently.
In a dataset, we always have a single working domain, which contains all the documents.
However, we can split these documents into multiple different categories, which are specific for this domain.
Therefore, it makes no sense to extract keywords which are specific to the domain, but frequent and important for each category (and probably most of the documents as well), since they do not help in differentiating between different documents and categories.

The important notice is that we do not focus on text categorisation itself but on the quality of document's representation that we could use for many tasks, including text categorisation.
A chosen word discriminative metric strongly influences the proposed method.
If the metric uses frequency statistics within categories, it can be used to create an efficient and highly discriminative representation of documents given the domain with multiple categories a priori.
By adding keyword metadata, it simplifies following tasks, but it is not able to categorise new unlabelled documents.
However, if we can compute the chosen discriminative metric from frequency statistics which is independent of category labels, it can be used to categorise new unlabelled documents as well.
We use text categorisation task to evaluate the quality of representation produced by our method, which allows us to demonstrate the extent of our contribution by comparing our results to the current state-of-the-art.

\section{Related work}
\label{sec-related-work}
Researchers have studied the extraction of relevant keywords in various domains.
Many of them focus on solving text categorisation problem by leveraging various discriminative word frequency statistics.
These are metrics or weighting schemes that weigh words by their importance, or more specifically in this case by their discriminativeness.
We can compute most of these metrics by using four variables $A$, $B$, $C$ and $D$ (see Table \ref{ABCD-table}, we refer to this later as the $ABCD$-statistics).
We can compute them as a sum over all documents in a corpus, given category CAT and word W (see \cite{Lan:2006,Park:2008,Debole:2003} for more details).

However, not all the metrics that we can express as a combination of these four variables are necessarily discriminative.
We can obtain the discriminative information from category frequency.
If we use only the sums of columns from Table \ref{ABCD-table}, namely $A + C$ and $B + D$, we are back to regular word frequencies, and we lose the discriminative information, since $A + C$ is equal to the frequency of word $W$ and $B + D$ is the frequency of all words except W.
However, if we combine these variables differently, if we apply a different function than summation to the pairs above or sum the variables across rows or diagonals, we can get a discriminative information.

\begin{table}[h]
	\caption{\label{ABCD-table} Frequency table given word W and category CAT. }
	\begin{center}
		\begin{tabular}{lcc}
			\toprule
			\bf Frequency & \bf of word W & \bf of other words \\
			\midrule
			\bf in category CAT & A & B \\
			\bf in other categories & C & D \\
			\bottomrule
		\end{tabular}
	\end{center}
\end{table}

The metrics utilising these four variables are summarised in Table \ref{metrics-table}, where N is sum of all these four variables (note that \textit{gr} metric uses \textit{ig} metric in its computation).

\cite{Ozgur:2005:TCC:2103951.2104023} also focus on class-based keyword selection, which is similar to computing a discriminative statistics.
However, they do not consider the inter-class relationships of word frequency statistics like the above-mentioned discriminative metrics do.
They compute \textit{tf-idf} metric for each category separately.

\begin{table}[h]
	\caption{\label{metrics-table} Frequency table given word W and category CAT. }
	\begin{center}
		\begin{tabular}{lc}
			\toprule
			\toprule
			\bf Metric name &
		\bf \begin{tabular}{@{}c@{}} Function expressed in terms \\ of \textit{ABCD}-statistics \end{tabular} \\
			\midrule
			\midrule rf (relevance frequency) & $log(2+\frac{A}{max(C,1)})$ \\
			\midrule tds (term domain specificity) & $\frac{A/(A+B)}{(A+C)/N}$ \\
			\midrule ig (information gain) &
			\begin{tabular}{@{}c@{}}
				$N \times $ \\
				$ \frac{A}{N} \times \frac{log(AN)}{(A+B) \times (A+C)} \times $ \\
				$ \frac{B}{N} \times \frac{log(BN)}{(B+A) \times (B+D)} \times $ \\
				$ \frac{C}{N} \times \frac{log(CN)}{(C+A) \times (C+D)} \times $ \\
				$ \frac{D}{N} \times \frac{log(DN)}{(D+B) \times (D+C)} $
			\end{tabular} \\
			\midrule gr (gain ratio) &
			\begin{tabular}{@{}c@{}}
				$ -ig / $ \\
				$ (\frac{A+B}{N} \times log(\frac{A+B}{N}) + $ \\
				$ \frac{C+D}{N} \times log(\frac{C+D}{N}) ) $
			\end{tabular} \\
			\midrule $\chi^2$ & $ \frac{N \times (AD-BC)^2}{(A+B)\times(C+D)\times(A+C)\times(B+D)} $ \\
			\midrule idf (inverse document frequency) & $log(\frac{N}{A+C})$ \\
			\bottomrule
			\bottomrule
		\end{tabular}
	\end{center}
\end{table}

There is also a group of metrics that use different statistical features.
For example, \cite{Lopes:2016:ETD:2899074.2899126} present \textit{tf-dcf} (term frequency - disjoint corpora frequency) metric, which uses geometric composition of term frequency in complementary categories as an alternative (see Formula \ref{metric-tf-dcf}).
In practice, it has a disadvantage of low precision when dealing with more categories and fixed precision of floating point numbers, since each additional category means adding another multiplier into the formula denominator, causing the numbers to become indistinguishably small in effect.
As we can see, \cite{Lopes:2016:ETD:2899074.2899126} use a dataset with only four categories in the evaluation, so we cannot observe such effect there.

\cite{Kit:2008} present $thd$ (termhood) metric, which uses term rank instead of term frequency (see Formula \ref{metric-thd}).
However, by using the rank, we lose information about the frequency gap sizes between successively ranked words, which can negatively influence the results in case of more categories and higher topical diversity of categories.

\begin{equation} tf\mhyphen dcf_{w,c} = \frac{tf_{t,c}}{\prod_{\forall d \in C} 1 + log(1 + tf_{t,d})}
	\label{metric-tf-dcf}
\end{equation}

\begin{equation} thd_{w,c} = \frac{r_{t,c}}{|V_c|} - \frac{r_{t,g}}{|V_g|}
	\label{metric-thd}
\end{equation}

All of the methods mentioned above have a common pitfall of being just statistical metrics that do not take into account word semantics.
There are also approaches that attempt to improve on pure statistical metrics by using some form of semantic information.
\cite{Luo:2011:STW:1994471.1994932} use WordNet in attempt to upgrade tf-idf weighting scheme by utilising relations between most probable word sense and category senses.
While their method performs better than \textit{tf-idf} on Reuters-21578 dataset, they report inferior results with 20-newsgroups dataset.
Doubtfulness of WordNet contribution to text processing is emphasized also by \cite{Celik:2013:comprehensive} when compared with \cite{Mansuy:2006:CWF:1273808.1273822}.
While the former reports superior results when using WordNet features and SVM classifier, the analysis presented in the latter reports that incorporating various WordNet features had no statistically significant positive effect on the accuracy of the Naive Bayes and SVM classifiers.

In the past, we also tried to use WordNet to improve on standard statistical methods that do not consider semantics.
Our previous approach \citep{Sajgalik:2013:ambiguous} extracts key-concepts (based on WordNet synsets) instead of words to classify documents using two-pass PageRank combined with concept probability.
With 20-newsgroups dataset, we reached the micro-F1 score of only $58.63\%$ using naïve Bayes classifier and $55.85\%$ using kNN classifier.

Apart from approaches based on statistical metrics, we can also find methods based on distributed representation.
\cite{Larochelle:2008} use Discriminative Restricted Boltzmann Machines to model document semantics.
The main difference from our approach is that they use binary vectors, not continuous.
They achieve $76.2\%$ micro-averaged F1 score on 20-newsgroups dataset.
\cite{Li:2010:IMT:2143642.2143648} propose an error-correcting output coding method, which also encodes documents into binary vectors.
Each bit in the document vector is computed in separate binary classification task using naïve Bayes classifier, where we partition all the categories into two super-classes, categorise the document and put the computed label as a value of the current bit in the document vector.
They achieve average accuracy of $81.84\%$ on the 20newsgroups dataset and $89.75\%$ on Reuters-21578 dataset.

There are also some approaches that utilise distributed representation of words and evaluate the discriminativeness of the proposed methods.
\cite{Marujo:2015:automatic} present one such example, where they combine Brown clusters with pre-learned continuous word vectors to enrich the feature set of their supervised keyword extraction method.
However, they report that adding word vectors does not improve the results very much.
We think that since they train vectors in an unsupervised manner, and the only tuned parameter is the number of extracted keywords, it is possible that \cite{Marujo:2015:automatic} overfit on other features due to random irregularities in the training data.
Therefore, the decision tree may be trained to ignore the latent features of word vectors most of the time.

As we use additional data like word2vec pre-trained vectors and Google N-gram Corpus in our method, we can consider it as a kind of semi-supervised method.
There are several semi-supervised methods which could be compared to ours.
\cite{Dai:2015:semi} use word2vec pre-trained vectors to train LSTM neural network and show how it helps to improve over purely supervised approaches.
They compare only neural network based approaches to showcase the improvement.
However, neither of the compared methods achieves state-of-the-art results in text categorisation.
We can find another example of a semi-supervised method in \cite{Zhang:2012:semi}, however, they only simulate the semi-supervised training by splitting the categorisation corpus into labelled and unlabelled examples.

To summarise, most of the statistical methods are not aware of word semantics.
The methods that use WordNet need to cope also with other non-trivial problems like word sense disambiguation and heuristic approximation of quantitative information about word relations.
Finally, the methods that use distributed representation of semantics operate only on document level or do not use the full potential of the latent features of word vectors.
In our approach, we attempt to overcome these shortcomings by embedding words and documents into a common semantic space with a combination of discriminative information to enhance the categorisation performance.

\section{Method description}

We propose a novel method to extract discriminative keywords of a document, which does not rely on any externally supplied human-labelled semantics such as WordNet ontology.
Instead, it uses distributed representation of words, which maps words into a multidimensional continuous feature space, where each word is represented as a vector of latent features that contain semantic and lexical information \citep{Mikolov:2013:linguistic}.
This brings us additional benefits such as an ability to measure the similarity of words (by employing cosine similarity of corresponding feature vectors) or to compute the representation of multi-word phrases using feature vector composition \citep{Mikolov:2013:distributed}.

Our method, based on these word feature vectors, consists of the following steps producing top-N discriminative words for a document:

\begin{enumerate}
	\item Extract candidate phrases from a document
	\item For each phrase compute its vector representation
	\item For each word in the model, calculate word relevancy based on some chosen word-weighting metric
	\item Sort words by their overall relevancy
\end{enumerate}

\subsection{Extracting candidate phrases}

\begin{figure}
	\centering
  \includegraphics[width=6cm]{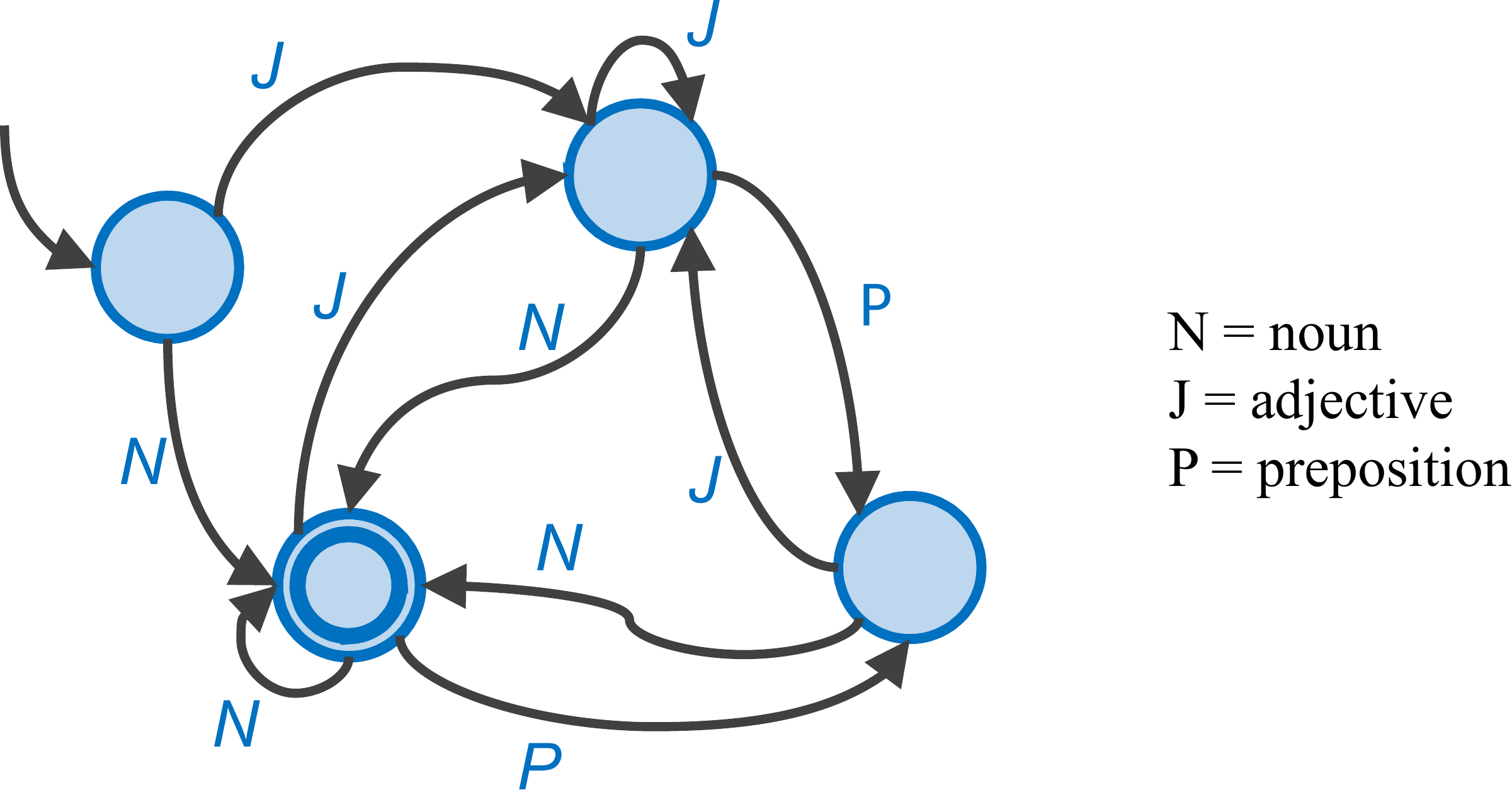}\\
  \caption{Finite-state machine to extract candidate phrases.}
  \label{fig-phrase-automaton}
\end{figure}

In this pre-processing step, we are interested in such phrases that have a potential to represent some meaningful concept bound to a topic discussed within a document.
Our approach is to extract phrases using a finite-state machine (see Figure \ref{fig-phrase-automaton}) operating on top of part-of-speech tags of words, being inspired by \cite{Vu:2008}\footnote{See \ref{appendix-pospatterns-phrases} for examples of extracted phrases}.
The state machine is designed as a simple and quick way to extract nouns as well as compound nouns (combination of adjectives, nouns and prepositions) since those can represent some reasonable concepts in the first place.

The main difference from \cite{Vu:2008} is that we limit the maximum length of extracted phrases to five words.
There are several reasons for doing so.
First, the longer phrases do not seem to have a significant impact on the results, since they are rarely used.
Second, it is computationally more efficient to have such limitation.
Lastly, the summing of vectors is most precise up to the maximum distance of words in the context.
According to \cite{Mikolov:2013:efficient}, this is set from 5 to 10.
Beyond that count, the sum of vectors becomes noisy since the model was not trained for that.

Another important difference is that we do not filter the generated phrases to be non-overlapping, which gives us more similar phrases.
However, since we care about text categorisation, where more features can help the classifier in better performance, we omit the relatively complicated approach in \cite{Vu:2008} to filter the extracted phrases while keeping our approach simpler.
After this step, we end up with a ``bag-of-phrases'' representation of a document, which means that each document is represented as a multiset of extracted phrases.

Next, we tried to augment this list of phrases with noun phrases obtained by parsing each sentence\footnote{We use Shift-reduce parser of Stanford CoreNLP.}.
After analysis of generated noun phrases we observed that the Stanford parser sometimes generates single-word phrases, where the word is not a real noun, but just a pronoun.
However, even after filtering out the pronouns, we did not observe any improvement compared to the automaton-based only approach mentioned above.

Note that the proposed state machine generates only phrases that always end with a noun.
It never generates phrases other than nouns and compound nouns.
Naturally, we also tried to extract other types of words, such as adjectives, verbs and adverbs.
Again, similarly to our previous attempt, extraction of additional types of words did not improve results compared to the approach based only on nouns.
Therefore, our method uses only the state machine to extract the multiset of candidate phrases.

\subsection{Transforming phrases into vectors}

As we already mentioned, representation of words based on feature vectors enables us (among other benefits) to sum multiple vectors of words to compose their joint meaning or to subtract vectors of words to obtain feature vector of their relation.

We utilise this advantageous property to compose meaning of extracted candidate phrases into the corresponding feature vectors.
The used feature vector model\footnote{
Available at \url{https://code.google.com/archive/p/word2vec/} in February 22, 2017
} contains not just words, but also many multi-word phrases that have a specific meaning different from what we would get if we have just summed up the vectors of words in those phrases.
We provide some examples of these multi-word phrases in Table \ref{table-w2vphrases} of \ref{appendix-w2vphrases}.

To compute a feature vector for given phrase, we need to split it into possibly multiple phrases which are present in the feature vector model.
Since we want to find the most reasonable interpretation of the split candidate phrase, we need to find such a mapping of the candidate phrases within our feature vector model, so that the number of splits of the candidate phrase is minimised and each split is present in the model.
In Algorithm \ref{algo-phrase_vector} we can see a reference pseudocode implementation by employing dynamic programming technique.
It is a simplified version of the algorithm used in \cite{Sajgalik:2014:exploring}, which also considered cases of ambiguous minimum length splits.
However, in practice, such ambiguous case is rare, so we prefer the simplified version.

\begin{algorithm}
	\caption{Algorithm for computing feature vector for a phrase.}
	\label{algo-phrase_vector}
Input: string \verb|PHRASE| of length \verb|N|

Output: feature vector of \verb|PHRASE|
	\begin{algorithmic}[1]
		\State Initialise array of integers \verb|NO_UNITS| of size \verb|N+1|, set all elements to \verb|0|
		\State Initialise array of vectors \verb|V| of size \verb|N+1|, set \verb|V[0]| to zero vector, the rest is undefined
		\For {$a \gets 0,$ \texttt{N}}
			\If {\texttt{V[a]} is not defined}
				\State continue
			\EndIf
			\For {$b \gets a+1,$ \texttt{N}}
				\If {\texttt{PHRASE[b]} is not end of word}
					\State continue
				\EndIf
				\If {\texttt{PHRASE[a:b]} is in the vector model}
					\State \verb|V_word| $\gets$ \Call{GetWordVector}{\texttt{PHRASE[a:b]}}
					\If {
						\texttt{V[b]} is undefined or\par
						\hskip\algorithmicindent~ \texttt{NO\_UNITS[a]} + 1 < \texttt{NO\_UNITS[b]} 
					}
						\State \verb|NO_UNITS[b]| $\gets$ \verb|NO_UNITS[a]| + 1
						\State \verb|V[b]| $\gets$ \Call{Normalise}{\texttt{V[a]} + \texttt{V\_word}}
					\EndIf
				\EndIf
			\EndFor
			\If {\texttt{PHRASE[a:b]} is not in the vector model} \label{code_lastif}
				\State \verb|b| $\gets$ \Call{GetEndOfNearestWord}{\texttt{PHRASE[a:]}}
				\State \verb|NO_UNITS[b]| $\gets$ \verb|NO_UNITS[a]| + 1
				\State \verb|V[b]| $\gets$ \verb|V[a]|
			\EndIf
		\EndFor
		\State \Return \verb|V[N]|
	\end{algorithmic}
\end{algorithm}

We use \verb|NO_UNITS| array to track the minimal number of unit phrases that comprise longer phrases.
Thus, we prefer unit phrases with multiple words, since their learned feature vector has higher quality than if we just summed up feature vectors of the individual words.
In case of ambiguity, we use \verb|SCORE| array to track the similarity of the composing unit phrases.
Thus, we prefer more common expressions.
The array V serves as an intermediate storage of computed vectors and is used to retrieve the final result.
The last conditional statement in line \ref{code_lastif} is a handler for unknown words (tokens) present in the phrase, which skips to the next available word.
This mostly handles the common cases of various punctuation characters present in the phrase which are not included in the model we used.

\subsection{Computing the set of discriminative words}

By generating a feature vector for each candidate phrase, we mapped the phrase into the multidimensional feature space, which means that we assigned a meaning to it.
However, having a vector of continuous features, it is hard to employ some frequency statistics.
Therefore we take word vectors which are closest to the phrase vector in the feature space within the external model of word vectors, effectively substituting each candidate phrase that was used to create the phrase vector by $K$ most similar words.
Substituting phrases with words simplifies further processing since we are back to words and we can easily utilise word frequency statistics in the next phase.
We can observe how working in the multidimensional feature vector space enabled us to overcome the problem of traditional statistical methods which do not recognise similar words like synonyms.

In fact, it is reasonable to consider two parameters - $K_1$ and $K_2$ - instead of the single parameter $K$.
When applying a statistical metric, we need to consider two input values - global word frequency and local word frequency.
The global word frequency represents word frequency across all documents, while the local word frequency represents the word frequency within a single document.
For example, in \textit{tf-idf} metric the \textit{tf} part is calculated of the local word frequency and the \textit{idf} part is calculated of global word frequency\footnote{
In contrast to the local word frequency which is equal to the word count, the global word frequency is always calculated as a word document frequency and is equal to the number of documents that the word appears in.
}.
Now we can consider different values of $K$ for calculating the global ($K_1$) and local ($K_2$) word frequencies.
However, in our experiments, we observed that the best results are always obtained with $K_1$ equal to $K_2$, although not exclusively.
Setting $K_2$ a bit bigger than $K_1$ makes no difference, but with $K_2$ lower than $K_1$ results in insufficient global information about the word, unbalanced with too much signal from local word frequency, thus effectively achieving worse results.
To summarise, the best strategy is to keep just one parameter $K = K_1 = K_2$, which simplifies the search for optimal parameter setting (the training phase) since we have fewer parameters to explore.

Extending each single phrase into tens of most similar words offers a better understanding of the meaning of a phrase, but also produces some noise, since not all of the words are equally important for given document.
To cope with this problem, we need to employ an importance metric to rank the words.
In text categorisation, we can equate the importance of a word to the ability of the word to discriminate between individual categories.
There are several such metrics (see Section \ref{sec-related-work}). It turns out that even after including the discriminating metric, we can further improve the quality of the extracted set of words.
After some observation, we can see that words obtained by substituting candidate phrases in the previous step are sometimes too specific.
That does not help in the case of sufficiently diverse categories which can be discriminated by more general terms.
Moreover, the weighting metric does not help us to solve this issue since it is applied only after the retrieval of the most similar words.
Therefore, we propose to crop the very model of feature vectors and get rid of the rare words.
We use Google N-gram corpus\footnote{Google N-gram corpus - available at \url{http://storage.googleapis.com/books/ngrams/books/datasetsv2.html} in February 22, 2017} to get word frequencies and retain only the most frequent words.
In our experiments, we use a fixed size of $20\,000$ most frequent words, which is approximately the size of a standard English dictionary.
Note that we still use the original model to calculate feature vectors for candidate phrases and we use the cropped version only afterwards in the computation of $K$ most similar words.

\section{Automatic evaluation of extracted keyword set}
We aim at evaluating the hypothesis that using our method we can extract a sufficiently discriminative set of words to be able to tell apart documents contained in different categories.
A natural implication is that upon succeeding, the extracted set of keywords for each document will not be the set of keywords that are important globally for given domain (like \textit{tf-idf}-like methods do), but discriminative words that are important for distinguishing a particular category.
In practice, we do not need words that are globally important, as we know the domain we are working in.
Since we sort the documents into the categories, those categories express the right level of topical granularity that interests us and words that have bigger discriminative power to categorise documents into such categories, are more informative and have bigger information value for us.

\subsection{Datasets}

We used four standard text categorisation datasets for evaluation.
We used 20-newsgroups dataset, which contains $18\,846$ Usenet articles (email messages) divided into $20$ categories \citep{Lang:1995:newsweeder}.
We used the version “by date” divided into training and testing sets, consisting of $11\,314$ and $7\,532$ documents respectively.
The second dataset was Reuters-21578, which contains $21\,578$ news articles annotated by topic labels \citep{Apte:1994:TLI:188490.188497}.
We used ModApte split, but also we filtered out all but documents with exactly one category, which resulted in having $6\,532$ and $2\,568$ documents in training and test set respectively, divided into $52$ categories.
An important thing to notice is that 20-newsgroups dataset has roughly equal number of documents per category, while Reuters-21578 has very skewed category distribution.

The other two evaluated datasets are WebKB and 7 Sectors.
They are both available from the WebKB project \citep{Craven:2000:LCK:350342.350347}.
Both these datasets differ from the previous two in several aspects.
First, they contain web pages as documents.
Second, the documents are organised into a hierarchy of categories.
Therefore, we preprocess the documents to obtain a textual representation of the content \footnote{
    We use $html2text$ Python package for that. Available at \url{https://github.com/aaronsw/html2text} in February 22, 2017
}.
We also use flattened versions of both datasets to avoid the problem of hierarchical categorisation.
In Appendix \ref{appendix-datasets}, we present distributions of documents over categories for each evaluated dataset.

To give an overview of the current state-of-the-art, the micro-F1 score for 20-newsgroups dataset achieves $84.86$ \citep{Rennie:2003:value}, for Reuters-21578 it is $94.66$ \citep{Suhil:2016:GRG}, for WebKB it is $94.1$ \citep{Sindhwani:2005:beyond}, and for 7 Sectors it is $88.02$ \citep{Erenel:2013:improving}.

\subsection{Methodology}

To evaluate the proposed method, we take the following steps:

\begin{enumerate}
    \item \label{item-transform} Transform list of extracted keywords to document feature vectors
    \item Search for optimal setup of the proposed method's parameters such as word-weighting metric or number of extracted keywords
    \item Run SVM classifier to categorise documents based on the feature vectors (step \ref{item-transform}) of respective documents
    \item Evaluate performance of categorisation
\end{enumerate}

\subsubsection{From keyword sets to document feature vectors}

We take advantage of having feature vectors for each word and simplify the representation of a document from a set of keywords to a single feature vector of a document.
This helps classifiers since it transforms the variable-sized set of keywords into a feature vector of fixed size, so we always have a reasonable number of input features to be used in training.
It helps mostly in cases when the cardinality of the extracted keyword set is small.
We also assume that similar documents have similar vectors that are closer to each other within a category and thus, classifiers can easily separate documents of different categories using such document feature vectors.
To transform the set of words into a feature vector, we simply sum up the feature vectors of each word in the extracted keyword set and normalise the resulting vector.
Thus, we end up with $300$ features represented by real numbers ($300$ is the dimensionality of vectors in the Google vector corpus we use) instead of $20\,000$ of binary features (depending on the vocabulary size) that we would otherwise work with if we used the keyword set.

\subsubsection{Parameter setup}

Since there are many parameters to be set in our method, it would be very time-consuming to evaluate every single combination on whole training set to choose the best one.
We need a smarter way of choosing the right values for them.
First, we need to set $K$, which tells us how many words we should use to enrich each phrase with.
Second, we need to choose the best metric for ranking the words.
We use a heuristic estimation of which combination of parameter values is the best.
We describe this heuristic in Algorithm \ref{algo-parameter-search}.

\begin{algorithm}
	\caption{Choosing the best performing setup of parameters.}
	\label{algo-parameter-search}
Input:

	$\quad data$ - input text documents

	$\quad setups$ - set of multiple combinations of parameter values to choose from

Output:

	$\quad$ - the best performing combination of parameter values

	\begin{algorithmic}[1]
		\State $categories \gets $ \Call{GetNumberOfCategories}{$data$}
		\For {$max\_doc\_count \in (5, 10, 20, 100, 1000) $}
			\State $cross\_validation\_ratios \gets (2, 3, 5)$
			\If {$cross\_validation\_ratio > 5$}
				\State $cross\_validation\_ratios \gets (5)$
			\EndIf
			\For {$cross\_validation\_ratio \in cross\_validation\_ratios$}
				\State $results \gets \varnothing$
				\State $best\_score \gets 0$
				\For {$setup \in setups$}
					\State $k \gets setup.k$
					\State $metric \gets setup.metric$
					\State $keywords \gets $ \Call{ComputeRankedKeywords}{$data, k, metric$}
					\State $keywords \gets $ \Call{Slice}{$keywords, setup.keywords$}
					\State $result \gets $

					\Call{ComputeKFoldCrossValidation}{$keywords, cross\_validation\_ratio$}
					\State $score \gets $ \Call{ComputeScore}{$result$} \label{code-compute_score}
					\If {$score > best\_score$}
						\State $best\_score \gets score$
					\EndIf
					\State $results \gets results \cup \{k, metric, keywords, score\}$
				\EndFor
				\State $best\_setups \gets \varnothing$
				\State $tolerance \gets \frac{cross\_validation\_ratio}{2 * max\_doc\_count * categories}$
				\For {$result \in results$}
					\If {$result.score >= best\_score - tolerance$}
						\State $best\_setups \gets best\_setups \cup result$
					\EndIf
				\EndFor
				\State $setups \gets best\_setups$
				\If {$|setups| = 1$}
					\State \Return \Call{GetAnyElement}{$setups$}
				\EndIf
			\EndFor
		\EndFor
		\State \Return \Call{GetAnyElement}{$setups$}
	\end{algorithmic}
\end{algorithm}

First, we gradually increase $max\_doc\_count$, which denotes the maximum number of documents that we take from each category.
The fewer documents we have, the faster is the SVM evaluation in \textsc{ComputeKFoldCrossValidation}.
We also found out that it is beneficial to try out multiple ratios in cross-validation.
However, we observed that it makes no practical difference whether we use it multiple times or just once.
Therefore we evaluate multiple cross-validation ratios just with the smallest $max\_doc\_count$ to filter out the most of bad performing setups right in the beginning.
After we run Algorithm \ref{algo-parameter-search} on whole training set, we use the obtained parameter setup to train the model on the same whole training set and to evaluate on the testing set to get the final score.

For function \textsc{ComputeScore} at line \ref{code-compute_score}, we use micro-averaged F1 measure (see Equation \ref{eq-microF1}) in all our experiments.
However, one is free to choose other measures according to the specific needs of working domain.
For example, one can be interested in better categorisation of documents only in more common categories and not the rare ones.
Alternatively, one can focus only on precision, while ignoring recall, etc.

Still, we need to account for the possibility of random noise in the obtained performance.
To avoid filtering out multiple good performing setups just because they miscategorised one or two documents, we set $tolerance$ variable to allow to retain such good setups.
Still, we experienced one anomaly that caused us to choose a wrong candidate setup.
In few cases, the methods with \textit{tf-chi-squared} and \textit{chi-squared} metrics were chosen as the best-performing ones since they outperformed other methods in early phases of Algorithm \ref{algo-parameter-search}.
However, the test results were poor for these methods.
We decided to filter out these two metrics from the candidate setups when also choosing the metric to get rid of this artefact, which worked well in all our experiments.

\subsubsection{Choice of a classifier}

We experimented with multiple different classifiers.
First, we tried $K$-nearest neighbours classifier, which seemed like a natural option for vector space we work with.
We tried using cosine distance as well as Euclidean distance as the distance metric. However, neither of them performed well.

We also tried linear discriminant analysis (LDA) presuming that each category can be expressed as a linear combination of features in a vector and thus not treating all features as equal could yield good results.
We found that from results obtained by LDA classifier we could approximate the results of SVM classifier quite well.

Most importantly, we tried linear SVM classifier, which we found having the best performance in other researchers’ work \citep{Lan:2006}.
However, it is not as straightforward as one might think.
Using different strategies for multi-class SVM yields different results.
For example, default strategy of multi-class SVM in libSVM implementation is using an ensemble of 1-against-1 two-class SVMs.
In this paper, we report results using 1-against-all strategy, which was substantially better, although it also took more time to train.
Based on our experiments, we suggest that it is best to use SVM classifier with the 1-against-all strategy for categorising documents represented as keywords extracted by our method.

\subsubsection{Measures of performance}

We evaluate the performance of our method using standard measures in text categorisation domain.
First, we use micro-averaged F1 score (see Equation \ref{eq-microF1}), which is computed of micro-averaged precision $P_{\mathrm{micro}}$ and recall $R_{\mathrm{micro}}$.
Second, we use macro-averaged F1 score (see Equation \ref{eq-macroF1}), which is computed of macro-averaged precision $P_{\mathrm{macro}}$ and recall $R_{\mathrm{macro}}$.
In both cases, $TP$ is the number of true positives, $FP$ is the number of false positives, $FN$ is the number of false negatives, and $C$ is the number of categories.

\begin{equation}
  \begin{gathered}
		P_{\mathrm{micro}} = \frac{\sum_{i=1}^{|C|}TP_{i}}{\sum_{i=1}^{|C|}TP_{i}+FP_{i}}
		\quad
		R_{\mathrm{micro}} = \frac{\sum_{i=1}^{|C|}TP_{i}}{\sum_{i=1}^{|C|}TP_{i}+FN_{i}}
		\\
		F_{\mathrm{micro}} = \frac{2P_{\mathrm{micro}}R_{\mathrm{micro}}}{P_{\mathrm{micro}}+R_{\mathrm{micro}}}
  \end{gathered}
	\label{eq-microF1}
\end{equation}

\begin{equation}
  \begin{gathered}
		P_{\mathrm{macro}} = \frac{1}{|C|}\sum_{i=1}^{|C|}\frac{TP_{i}}{TP_{i}+FP_{i}}
		\quad
		R_{\mathrm{macro}}=\frac{1}{|C|}\sum_{i=1}^{|C|}\frac{TP_{i}}{TP_{i}+FN_{i}}
		\\
		F_{\mathrm{macro}} = \frac{2P_{\mathrm{macro}}R_{\mathrm{macro}}}{P_{\mathrm{macro}}+R_{\mathrm{macro}}}
  \end{gathered}
	\label{eq-macroF1}
\end{equation}

\subsection{Results and discussion}

Here, we consider two convenient scenarios:

\begin{enumerate}
    \item We give an overview of how different statistical metrics used in word ranking influence the overall performance of our method.
    \item One can be interested in trade-offs of using a different number of keywords.
        For example, if we want to present extracted keywords to a user, we might prefer having fewer keywords.
\end{enumerate}

\subsubsection{Comparison of influence of different word-weighting metrics}

\begin{figure}[htb]
	\centering
	\resizebox{0.95\textwidth}{!}{%
		\includegraphics[height=1cm]{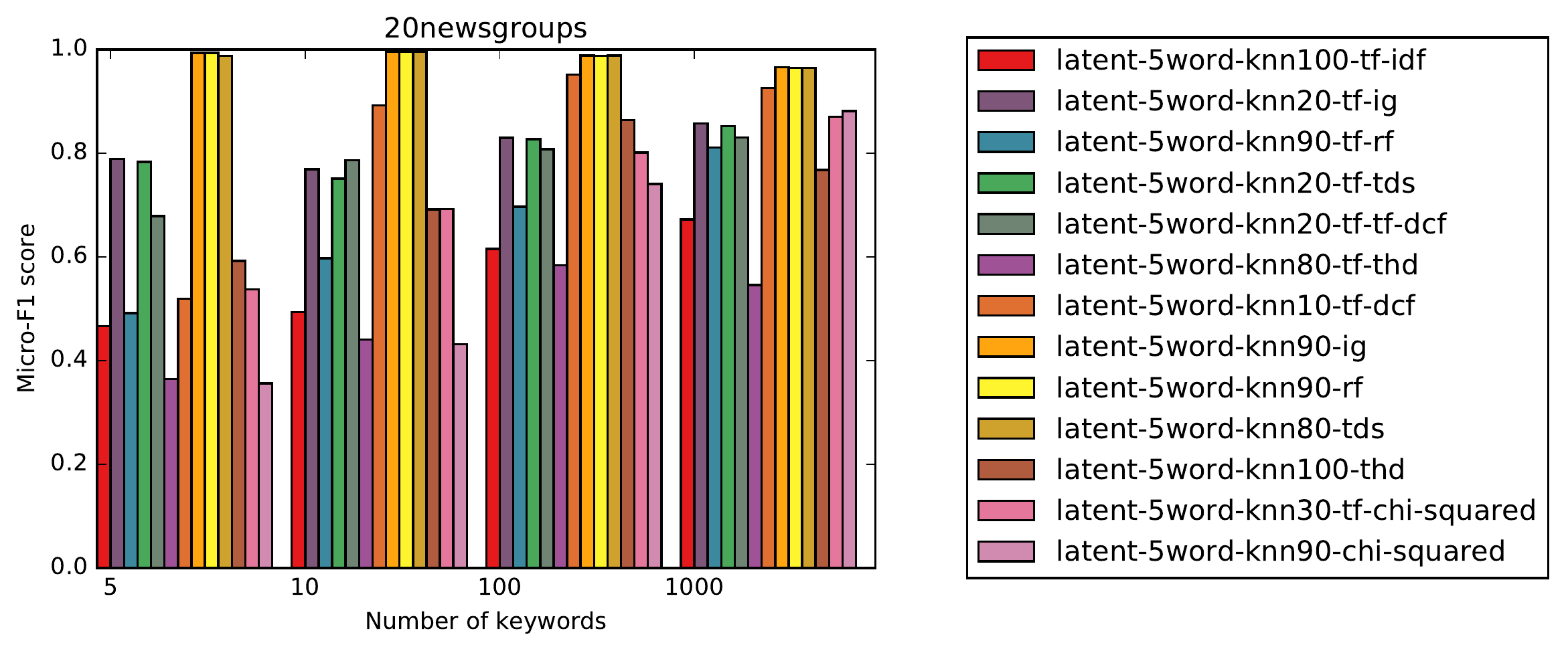}
		\includegraphics[height=1cm]{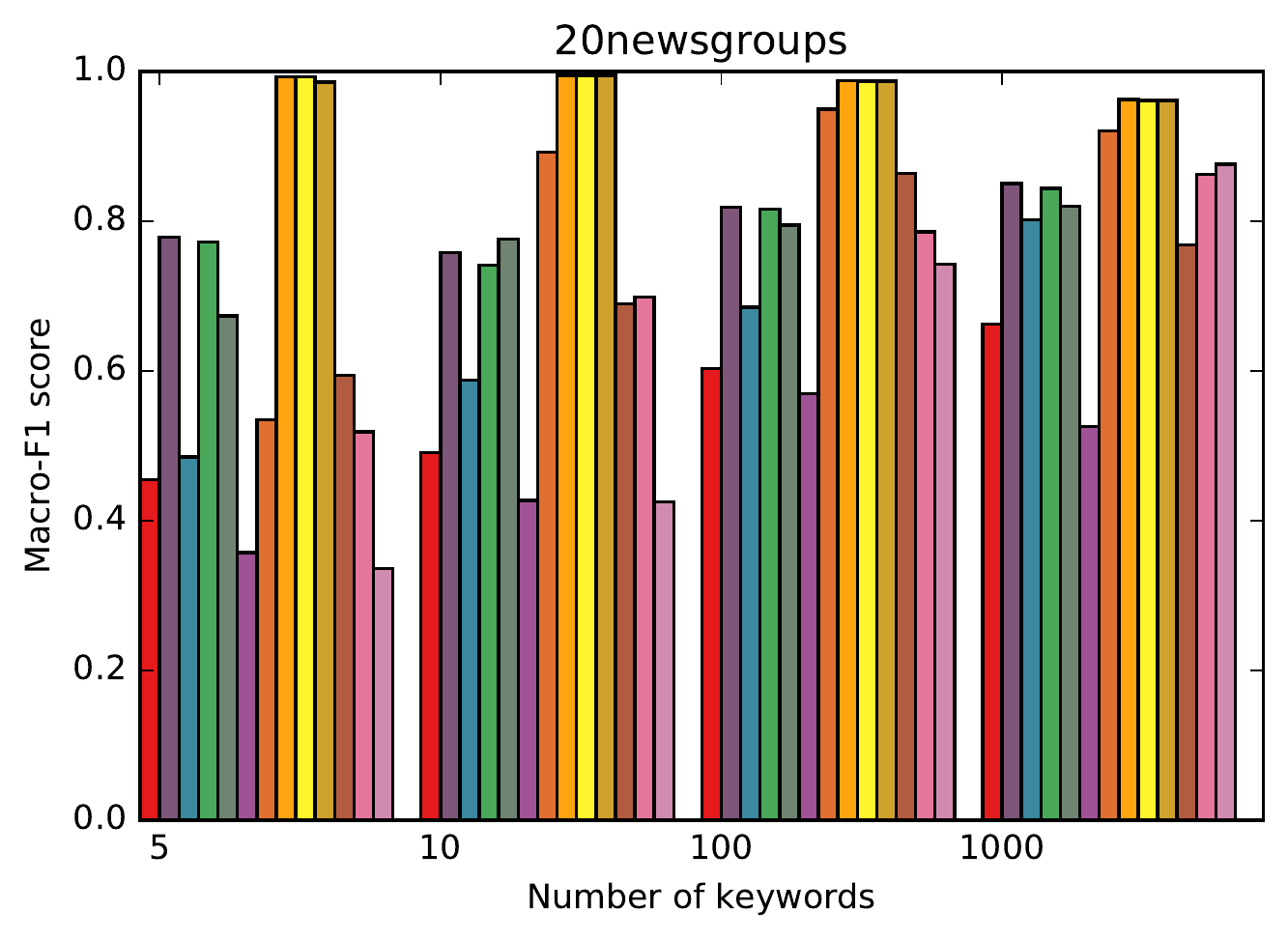}
	}
	\caption{Comparison of different statistical metrics used in ranking words evaluated on 20newsgroups dataset.}
	\label{fig-20newsgroups-f1-metrics}
\end{figure}

\begin{figure}[htb]
	\centering
	\resizebox{0.95\textwidth}{!}{%
		\includegraphics[height=1cm]{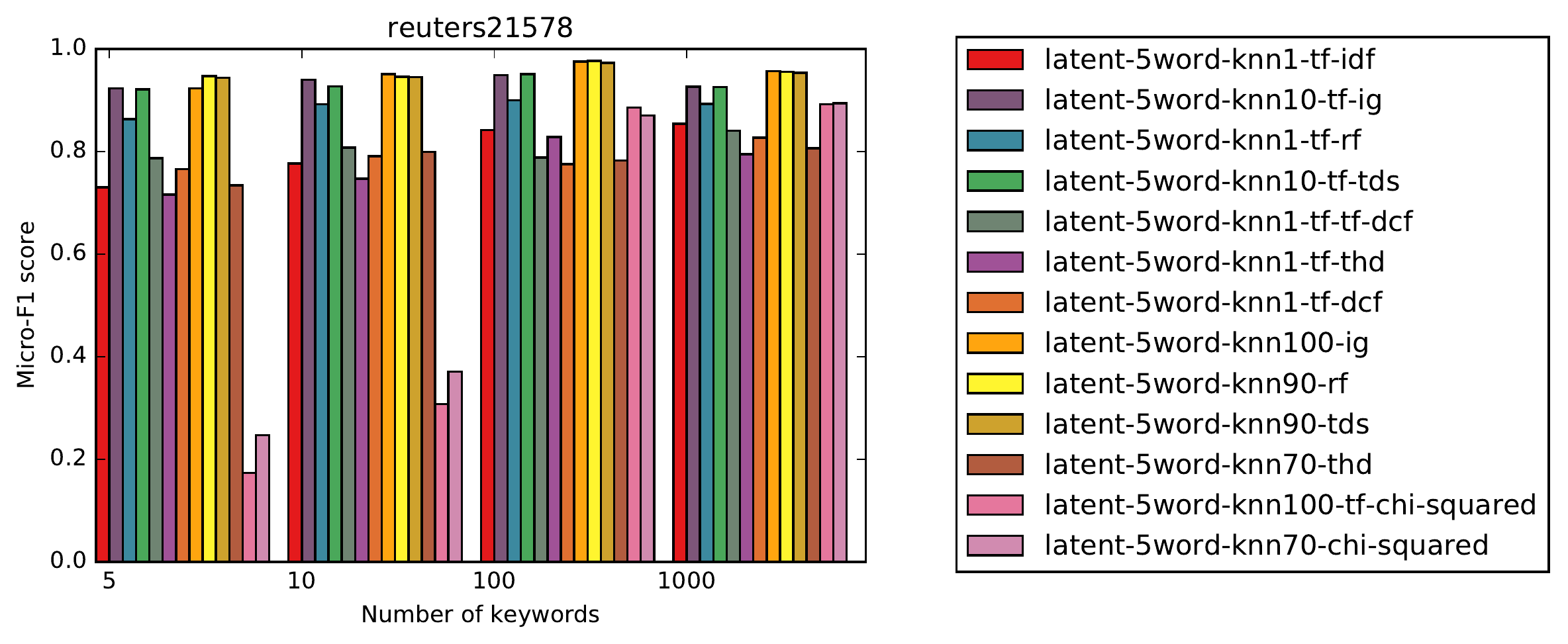}
		\includegraphics[height=1cm]{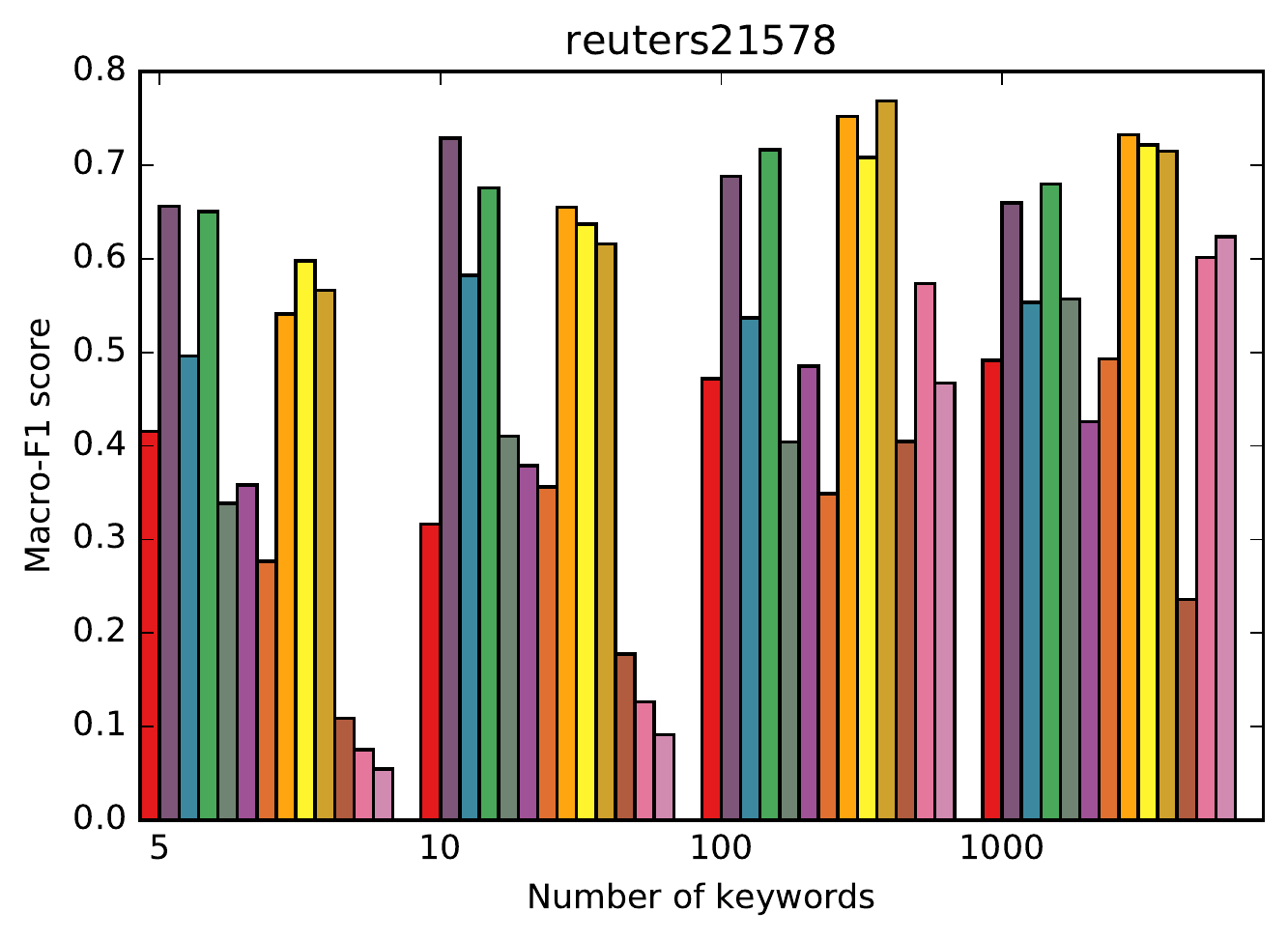}
	}
	\caption{Comparison of different statistical metrics used in ranking words evaluated on Reuters-21578 dataset.}
	\label{fig-reuters21578-f1-metrics}
\end{figure}

\begin{figure}[htb]
	\centering
	\resizebox{0.95\textwidth}{!}{%
		\includegraphics[height=1cm]{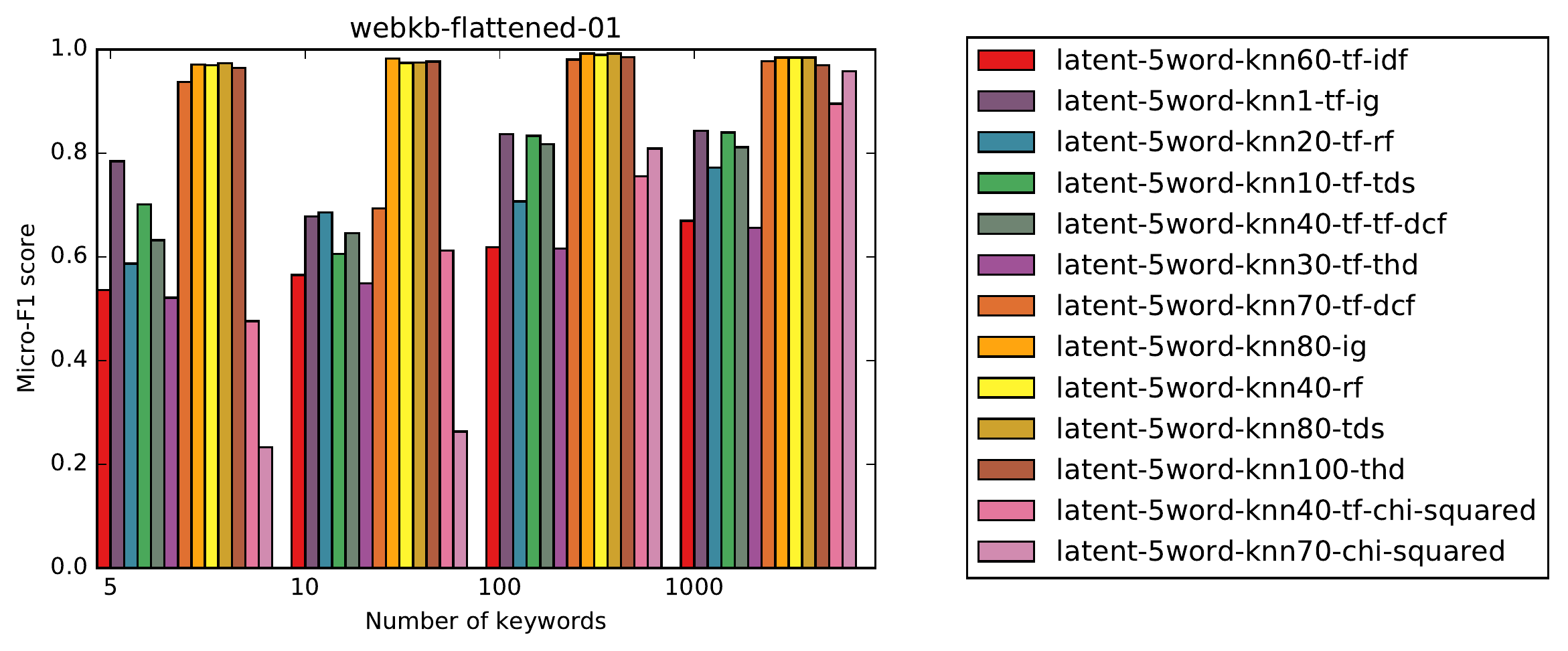}
		\includegraphics[height=1cm]{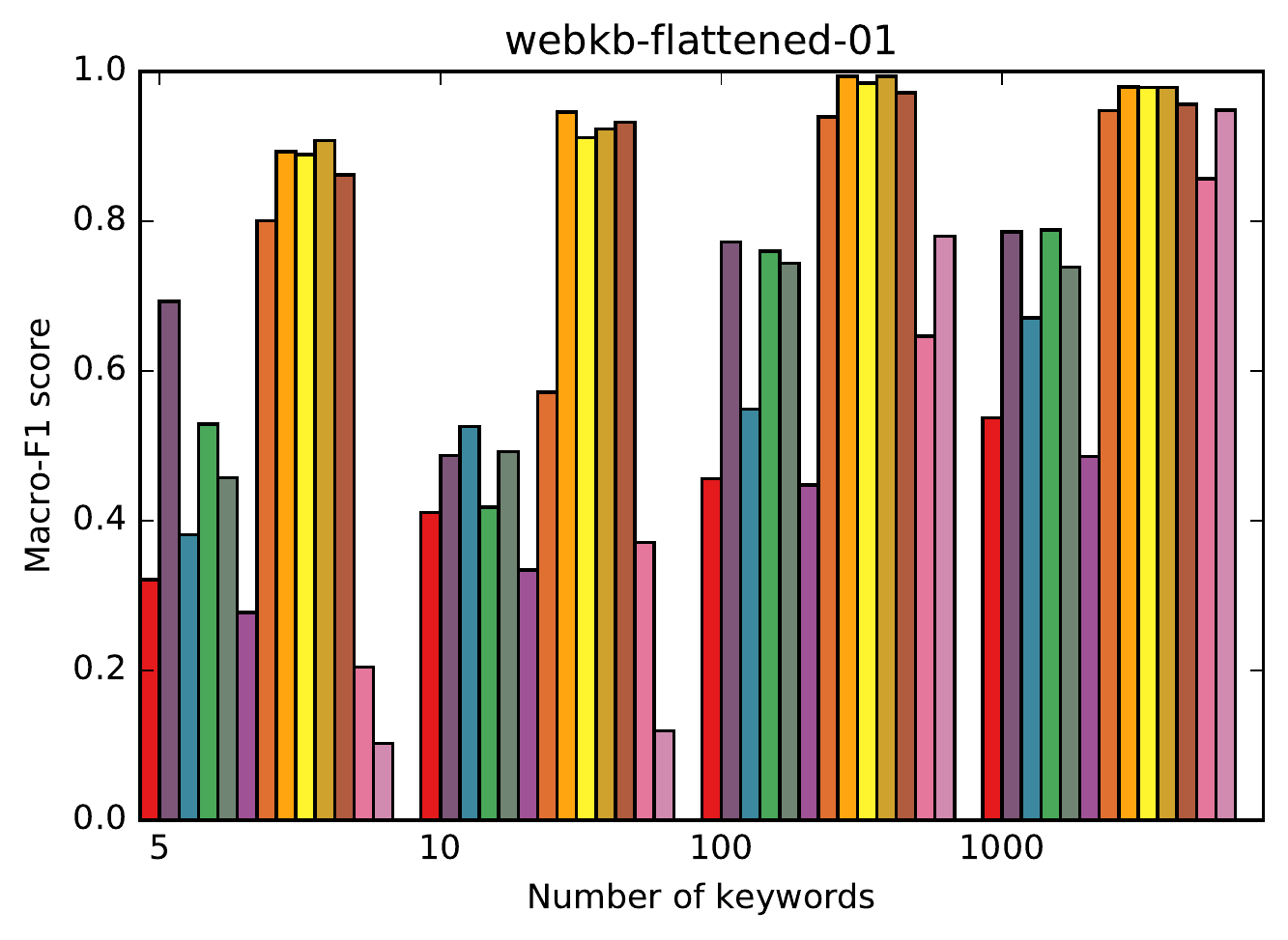}
	}
	\caption{Comparison of different statistical metrics used in ranking words evaluated on webkb-flattened-01 dataset.}
	\label{fig-webkb-flattened-01-f1-metrics}
\end{figure}

\begin{figure}[htb]
	\centering
	\resizebox{0.95\textwidth}{!}{%
		\includegraphics[height=1cm]{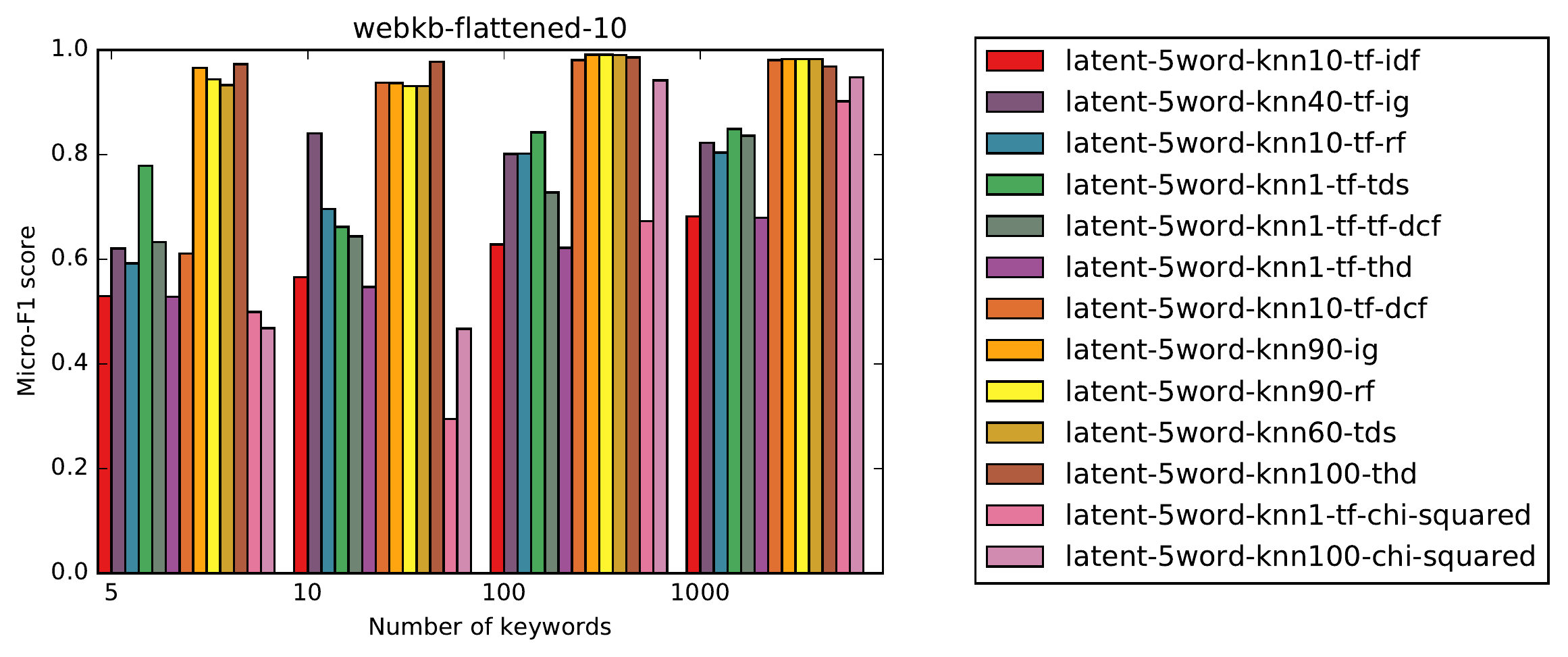}
		\includegraphics[height=1cm]{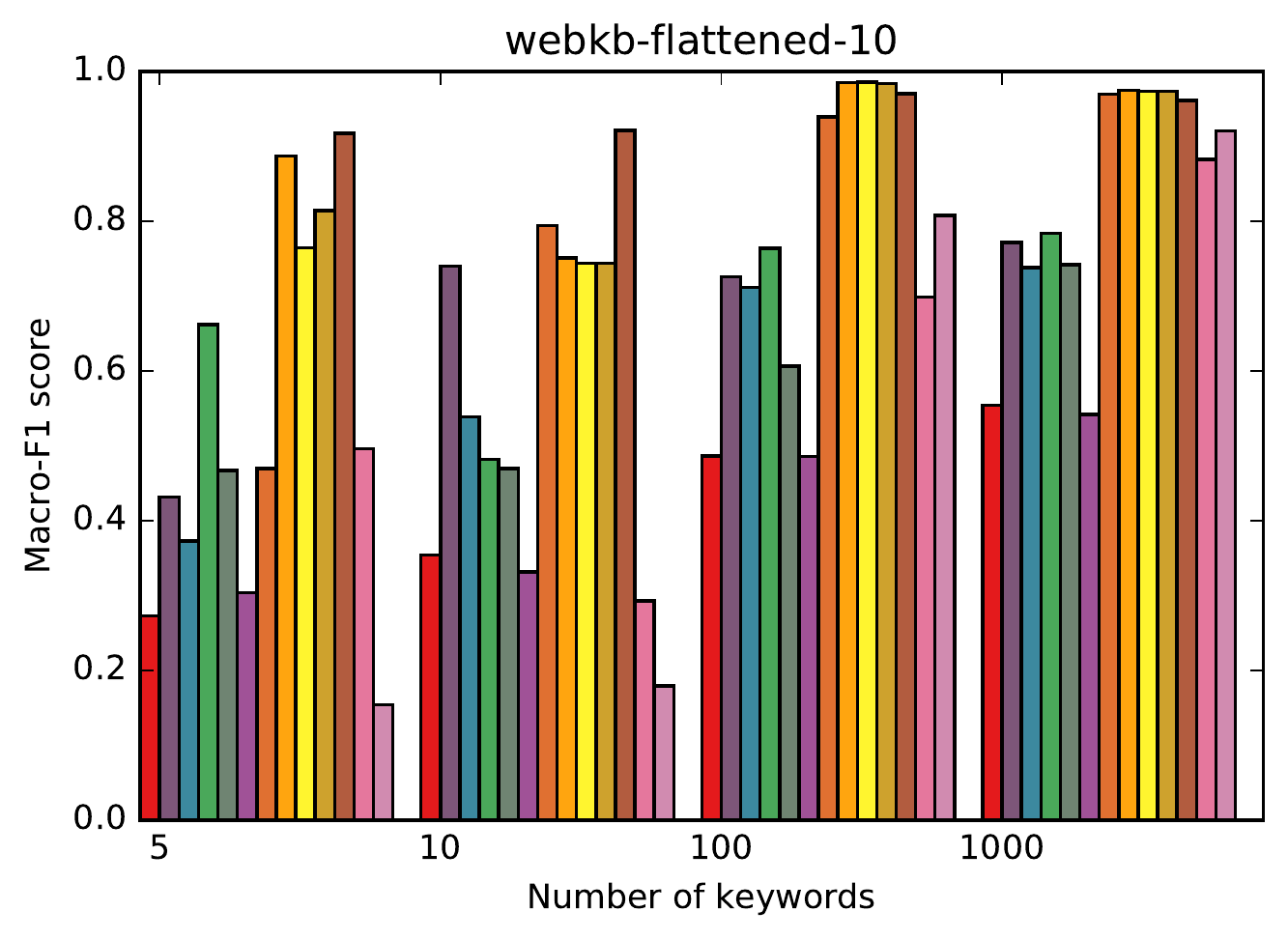}
	}
	\caption{Comparison of different statistical metrics used in ranking words evaluated on webkb-flattened-10 dataset.}
	\label{fig-webkb-flattened-10-f1-metrics}
\end{figure}

\begin{figure}[htb]
	\centering
	\resizebox{0.95\textwidth}{!}{%
		\includegraphics[height=1cm]{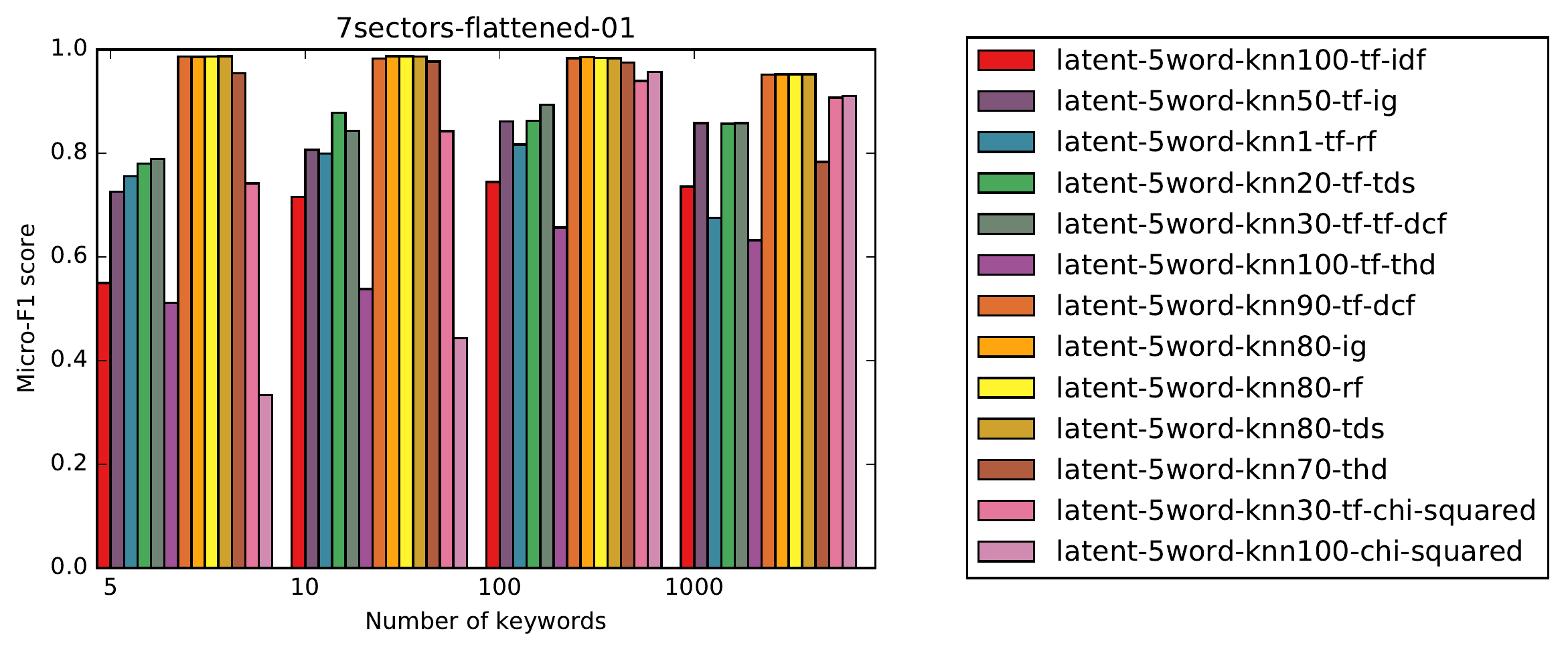}
		\includegraphics[height=1cm]{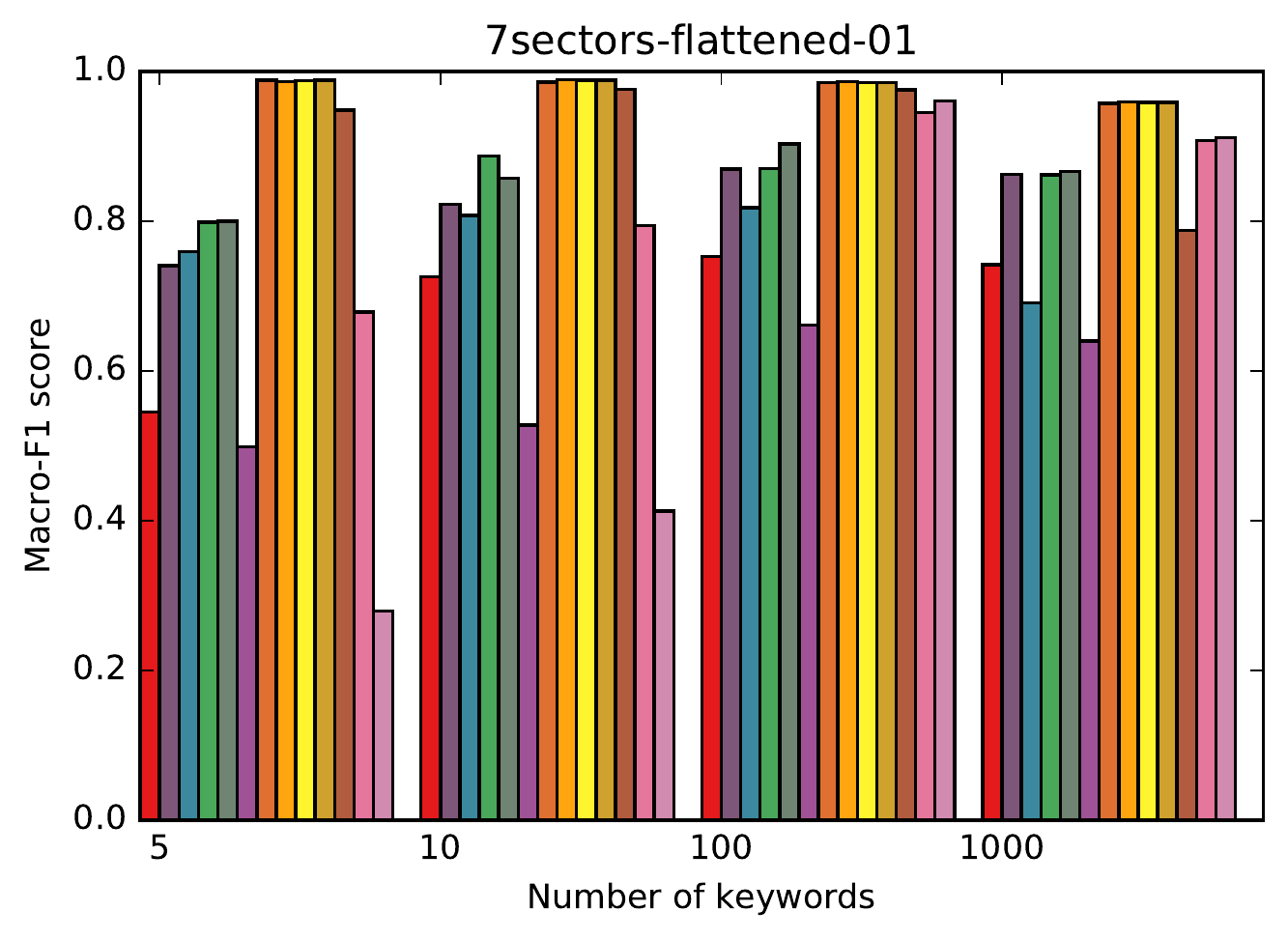}
	}
	\caption{Comparison of different statistical metrics used in ranking words evaluated on 7sectors-flattened-01 dataset.}
	\label{fig-7sectors-flattened-01-f1-metrics}
\end{figure}

\begin{figure}[htb]
	\centering
	\resizebox{0.95\textwidth}{!}{%
		\includegraphics[height=1cm]{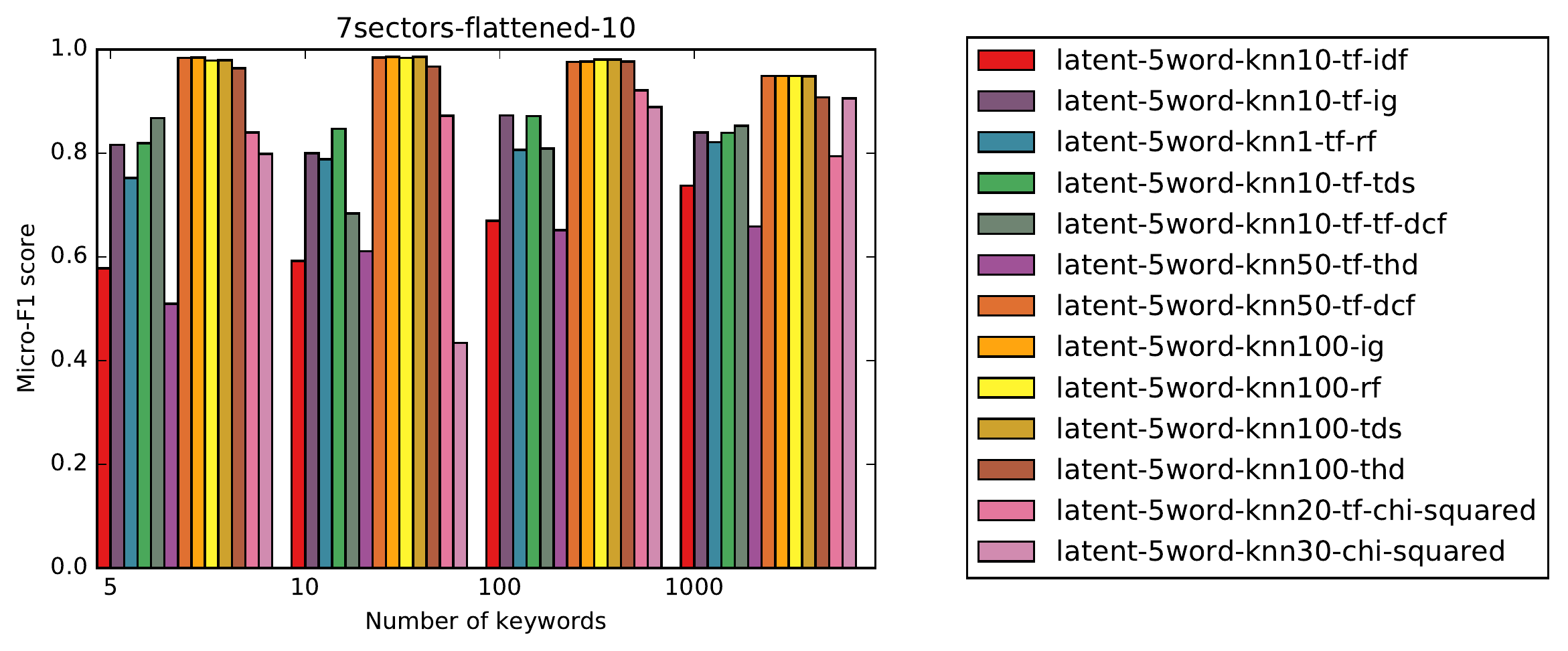}
		\includegraphics[height=1cm]{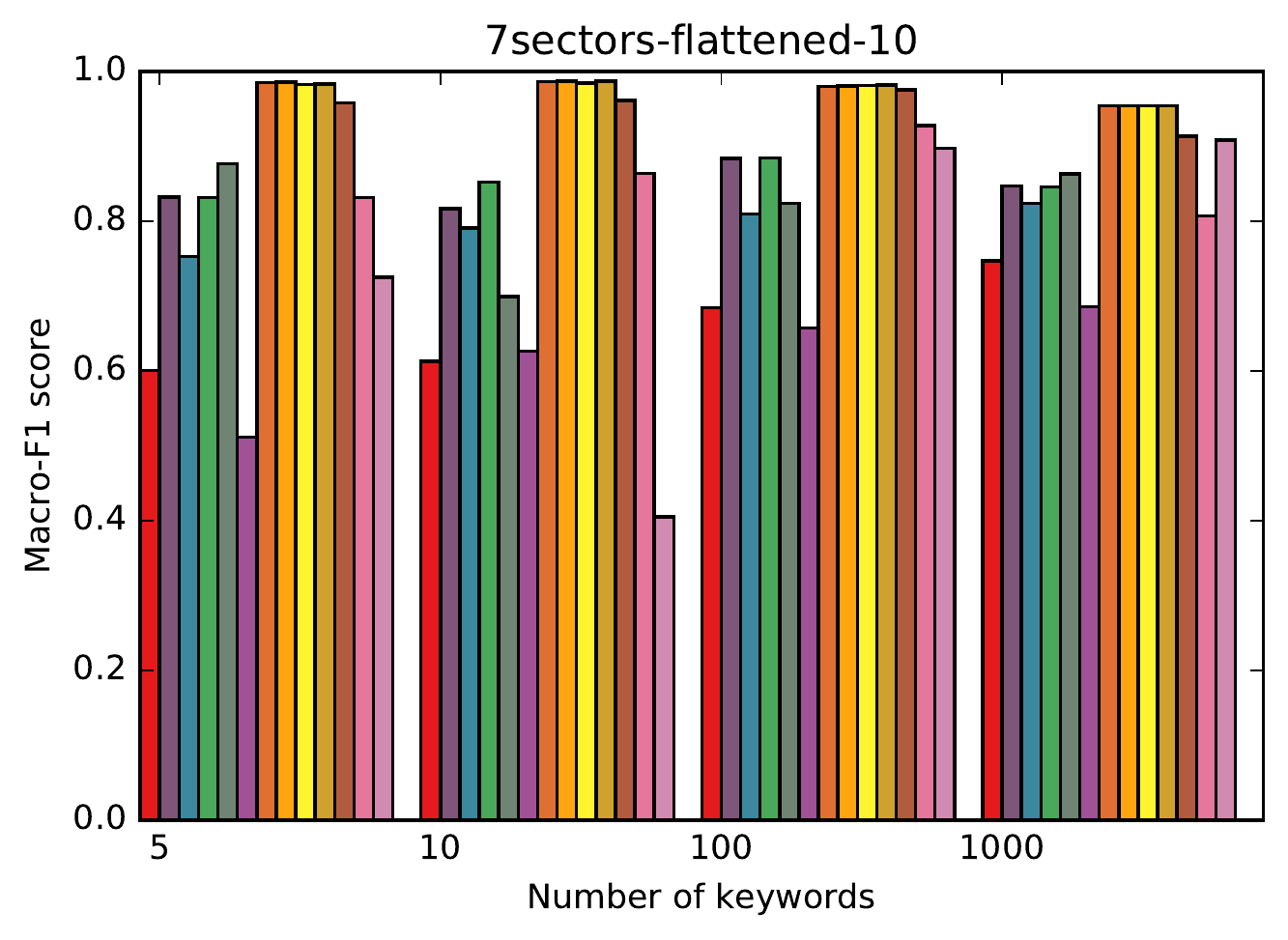}
	}
	\caption{Comparison of different statistical metrics used in ranking words evaluated on 7sectors-flattened-10 dataset.}
	\label{fig-7sectors-flattened-10-f1-metrics}
\end{figure}

In Figures \ref{fig-20newsgroups-f1-metrics} to \ref{fig-7sectors-flattened-10-f1-metrics} we can see how different metrics influence the overall categorisation performance.
For each metric, we used Algorithm \ref{algo-parameter-search} to find the best value for parameter $K$.
Starting with Figure \ref{fig-20newsgroups-f1-metrics}, we can see that methods with \textit{ig, rf} and \textit{tds} metrics are always among the best performing.
Another regularity that holds for all datasets is that methods with \textit{tf-chi-squared} and \textit{chi-squared} metrics are steadily improving with increasing number of extracted keywords.
Since 20newsgroups dataset has roughly balanced categories, we can see that there is almost no difference between micro-averaged and macro-averaged F1 score.

However, things get interesting when we look at the results for the Reuters-21578 dataset.
The Reuters-21578 dataset contains 52 categories, which is pretty much in comparison to other evaluated datasets.
Moreover, the distribution of documents over categories in the Reuters-21578 dataset is much skewed.
Almost 70\% of all documents are contained in only two categories.
More categories mean more similar categories, which requires better distinction of smaller grained topics.
We can see that methods with metrics containing the \textit{tf} factor have relatively better performance, since the \textit{tf} factor tends to propagate just the smaller grained local topics.
By definition, the \textit{tf} factor boosts words which are local to a document, unaware of word probabilities outside the document.
In results for the Reuters-21578 dataset, we can also see the clear difference between micro-averaged and macro-averaged F1 score.
Macro-averaged F1 score better presents categorisation performance on low-populated categories \citep{Sokolova:2009:SAP:1542545.1542682}.

Based on this information, we can see that very good micro-averaged F1 score of the majority of the methods for the Reuters-21578 dataset is caused mainly by its skewed distribution.
All those metrics can easily learn that there are two big categories forming 70\% of the dataset and the probability of a document belonging to them is therefore very high.
Since this skewed distribution is the same in both training and testing set, we need to check the macro-averaged F1 scores, which give us a much better impression of what is the actual difference between used metrics.

Looking at the rest, we can spot the difference between webkb-flattened (Figure \ref{fig-webkb-flattened-01-f1-metrics} and \ref{fig-webkb-flattened-10-f1-metrics}) and 7sectors-flattened datasets (Figure \ref{fig-7sectors-flattened-01-f1-metrics} and \ref{fig-7sectors-flattened-10-f1-metrics}).
Both have a small number of categories, but webkb-flattened is less diverse in topics since the flattened sub-categories represent semantically similar documents coming from different universities.
The hard part of correctly categorising documents in the webkb-flattened dataset is that the biggest category is \textit{other}, but there are still several other categories which are big enough so that we cannot rely on category probability in this case as we do in case of the Reuters-21578 dataset.
In contrast to other categories, the \textit{other} category contains more semantically diverse topics.

In the 7sectors-flattened dataset, the topical diversity of the flattened sub-categories is more balanced.
Moreover, the 7sectors-flattened dataset is less skewed in document distribution over categories than webkb-flattened.

We also experimented with using \textit{gr} metric for word-weighting (see Section \ref{sec-related-work}).
However, in all our experiments the proposed method with \textit{gr} metric gives the same results as with \textit{ig} metric.
In fact, there is a very slight difference between the two, which is barely observable, since the floating point number precision is limited.
It can be explained by the fact that \textit{gr} metric is computed out of \textit{ig} metric.
Thus, we omit results of our method with \textit{gr} metric, as we want to present more comprehensive results.

From results achieved on all datasets, we can conclude that the choice of the word-weighting metric is crucial.
By analysing the functions of different word-weighting metrics, we can conclude, that the best performing methods are using metrics based on $ABCD$-discriminative statistics.
We can see that methods with metrics like \textit{ig, rf} and \textit{tds} give roughly the best results on all datasets.
Other discriminative metrics like \textit{tf-dcf} and \textit{thd} can in some cases help our method to perform satisfiably on text categorisation task.
In particular, the methods with \textit{tf-dcf} metric seem to yield competitive results when the topical diversity of documents and the size of categories is well balanced (see Figure \ref{fig-7sectors-flattened-01-f1-metrics} and \ref{fig-7sectors-flattened-10-f1-metrics}), although we might need to extract more keywords when they are not quite that balanced (see Figure \ref{fig-20newsgroups-f1-metrics}, \ref{fig-webkb-flattened-01-f1-metrics} and \ref{fig-webkb-flattened-10-f1-metrics}).
From results in Figures \ref{fig-webkb-flattened-01-f1-metrics}-\ref{fig-7sectors-flattened-10-f1-metrics} we can see that methods with the rank-based \textit{thd} metric seem to favour small number of categories, since they give results comparable to the best performing methods with frequency-based metrics \textit{ig, rf} and \textit{tds}.
It also seems to give worse results when some of the categories are topically very similar to each other like in Figure \ref{fig-20newsgroups-f1-metrics}.
For methods with \textit{tf-chi-squared} and \textit{chi-squared} metrics we need to extract more keywords.
Although the methods with \textit{tf-chi-squared} metric can yield satisfiable results if we have a small number of topically well-balanced categories (see Figure \ref{fig-7sectors-flattened-01-f1-metrics} and \ref{fig-7sectors-flattened-10-f1-metrics}), it is impractically limited to just one ideal scenario while still not competitive with our best-performing methods.
As an interesting side point, we can see how the inner nature of \textit{tf-idf} (based on global word probability) helps our method to categorise documents correctly when the documents are coming from the most populated category (see the micro-averaged F1 score in Figure \ref{fig-reuters21578-f1-metrics}).

In \ref{appendix-metrics-significance} we provide complete results of statistical significance testing for each pair of metrics which we considered in our method.

\subsubsection{Comparison of achieved results for different number of extracted keywords}

Here, we compare results of the proposed method for the different number of extracted keywords.
For each number of extracted keywords, we first search for optimal setting of parameter $K$ and most suitable choice of a word-weighting metric with Algorithm \ref{algo-parameter-search}.

\begin{figure}[htb]
	\centering
	\begin{subfigure}[htb]{0.49\textwidth}
		\includegraphics[width=0.95\textwidth]{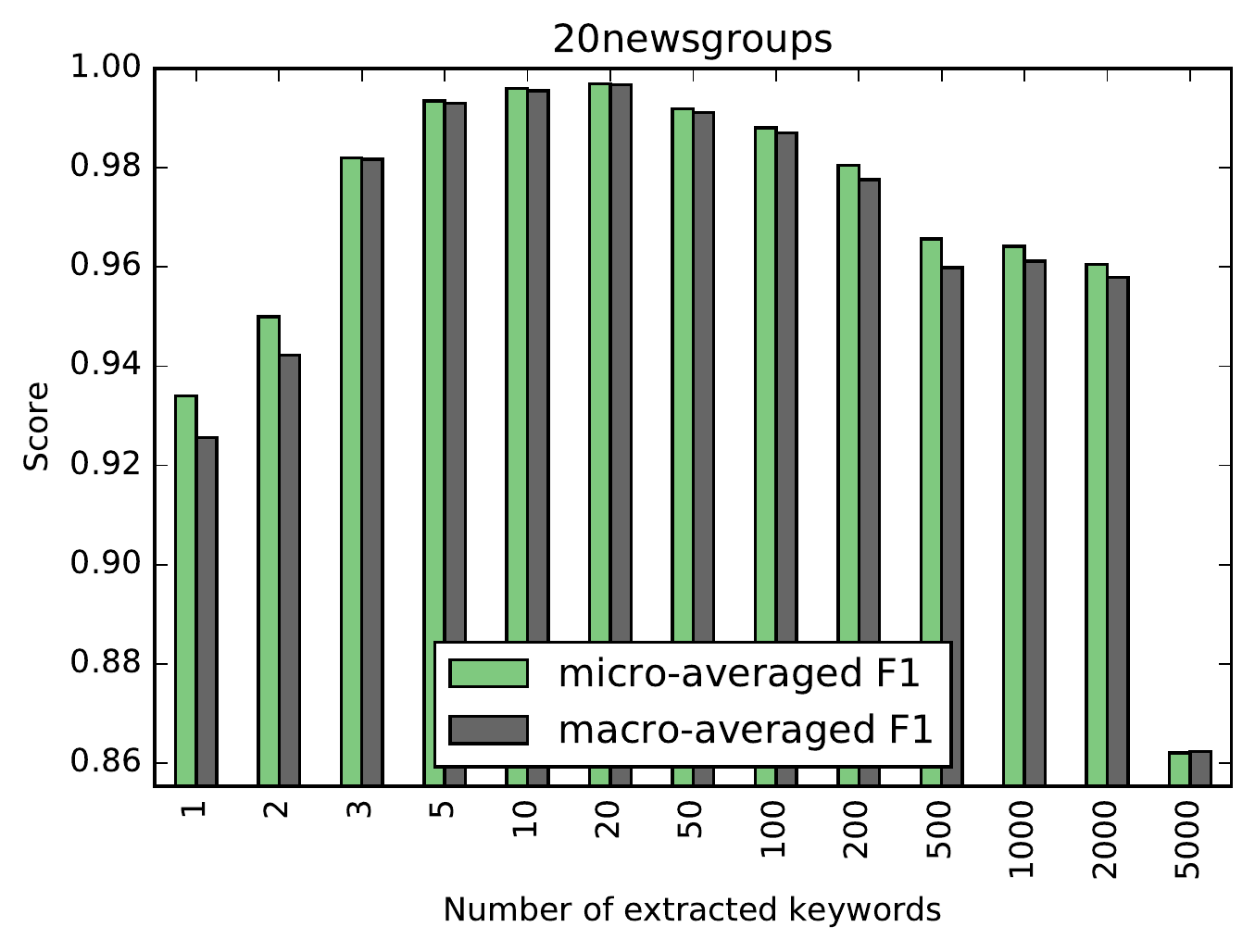}
		\caption{Evaluated on 20newsgroups dataset.}
		\label{fig-20newsgroups-keywords}
	\end{subfigure}
	\begin{subfigure}[htb]{0.49\textwidth}
		\includegraphics[width=0.95\textwidth]{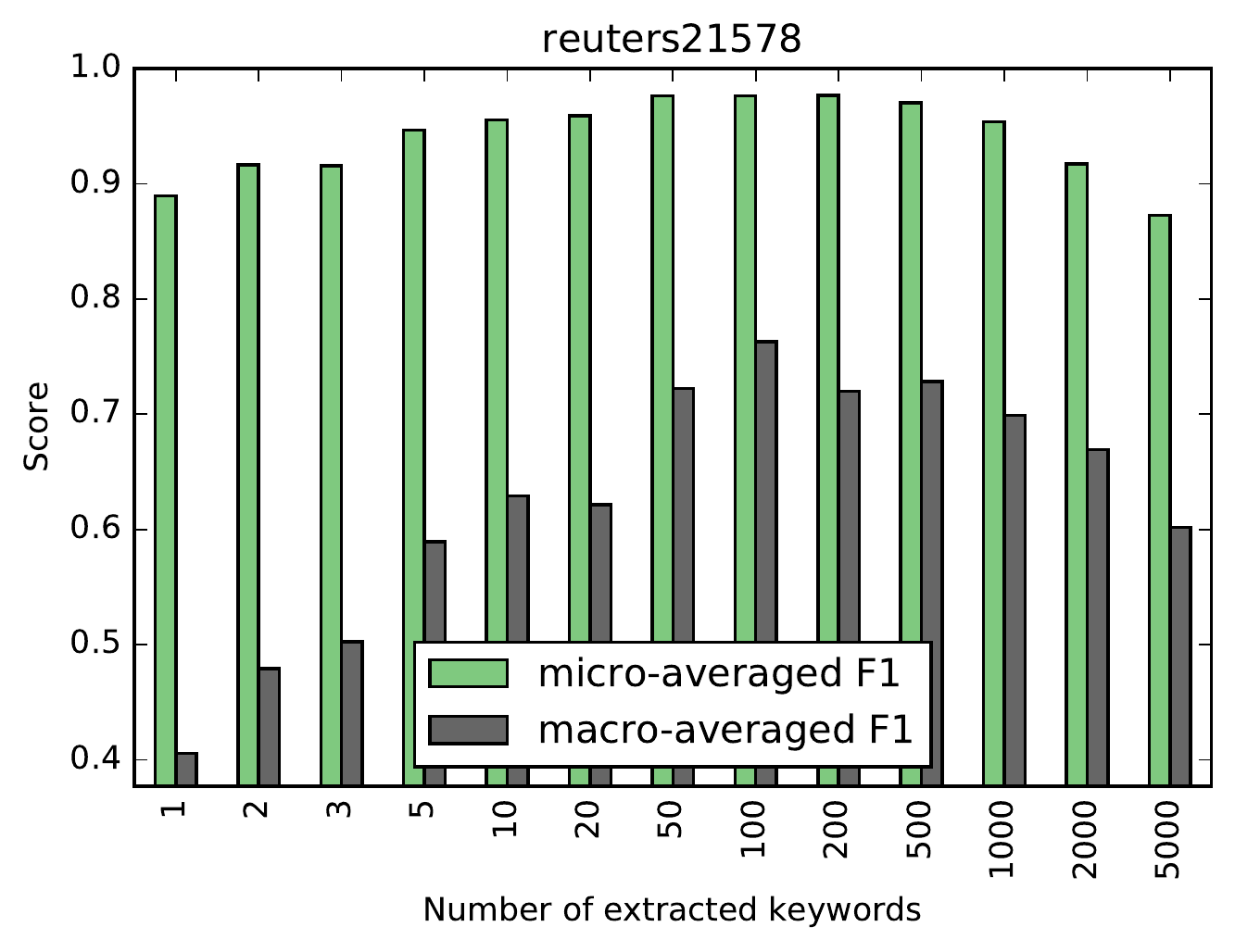}
		\caption{Evaluated on Reuters-21578 dataset.}
		\label{fig-reuters21578-keywords}
	\end{subfigure}
	\caption{Comparison of proposed method's results for different number of extracted keywords.}
\end{figure}

We can observe similar outcomes as in the comparison of the influence of different word-weighting metrics.
We can see that for the 20newsgroups dataset (Figure \ref{fig-20newsgroups-keywords}), both micro-averaged as well as macro-averaged scores share similar characteristics of being relatively stable having slightly convex progression with increasing number of extracted keywords.

\begin{figure}[htb]
	\centering
	\begin{subfigure}[htb]{0.49\textwidth}
		\includegraphics[width=0.95\textwidth]{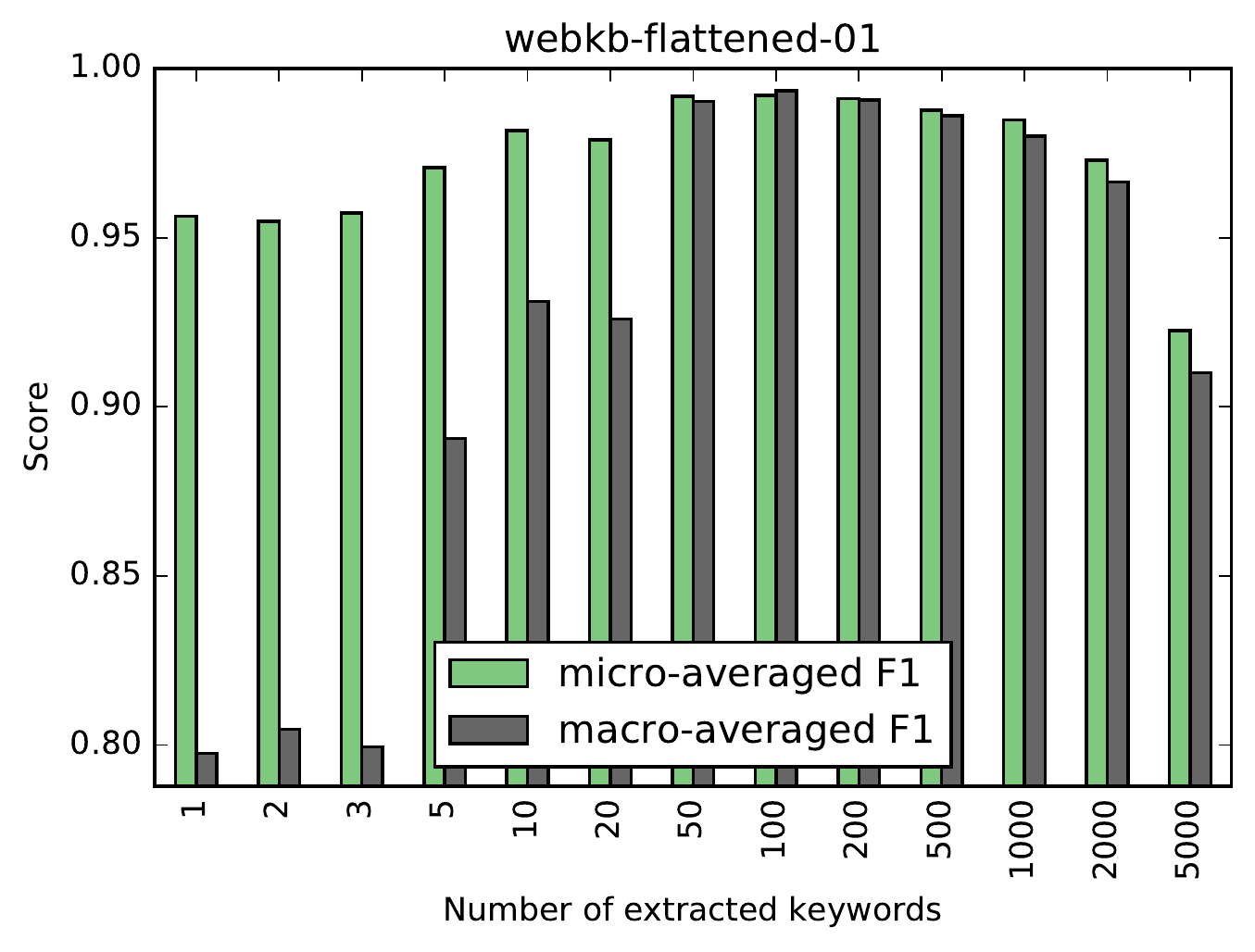}
	\end{subfigure}
	\begin{subfigure}[htb]{0.49\textwidth}
		\includegraphics[width=0.95\textwidth]{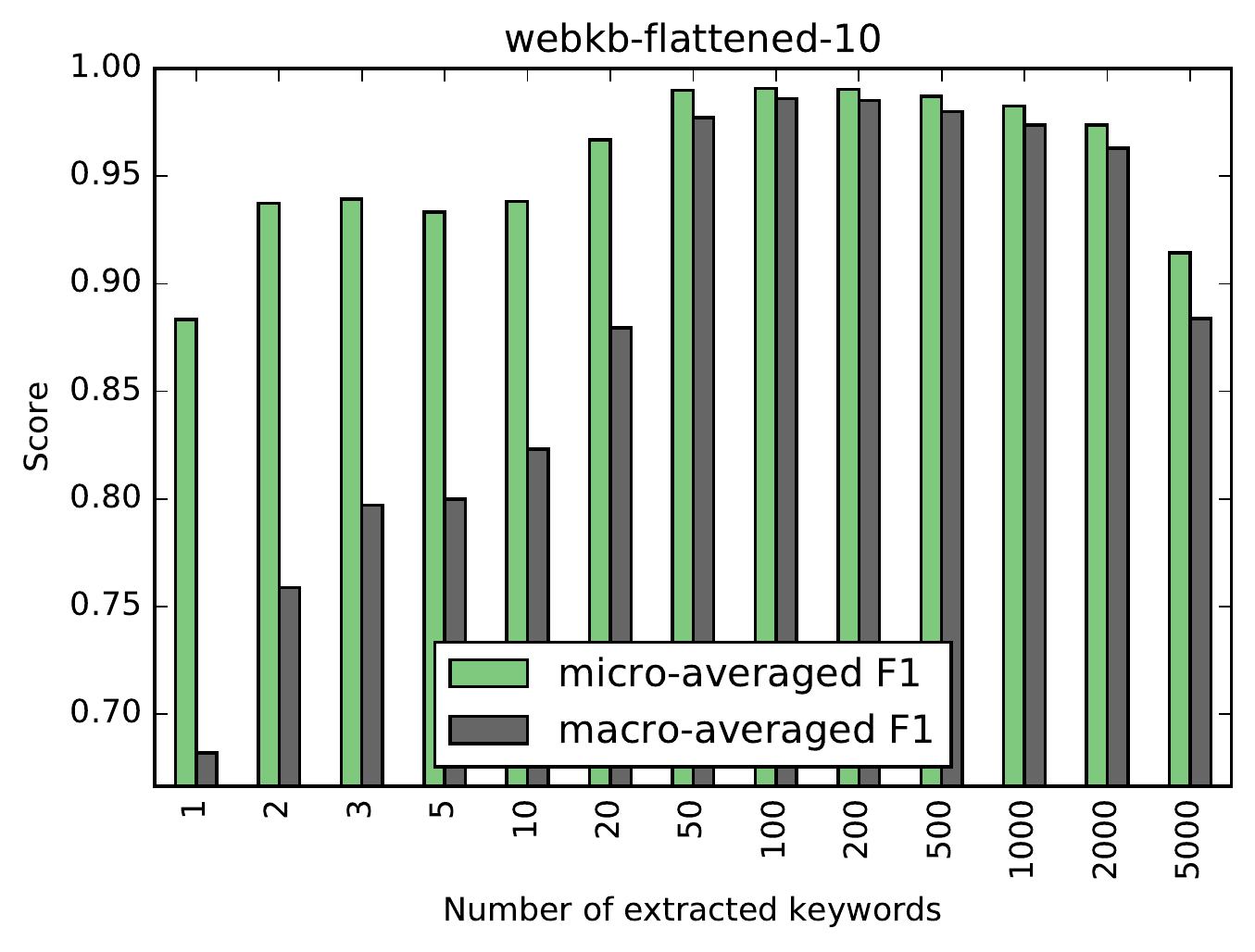}
	\end{subfigure}
	\caption{Comparison of proposed method's results for different number of extracted keywords evaluated on webkb-flattened dataset.}
	\label{fig-webkb-flattened-keywords}
\end{figure}

For Reuters-21578 dataset (Figure \ref{fig-reuters21578-keywords}), we can see that macro-averaged F1 score is much lower due to the high skewness of document distribution over categories.
We can observe similar impact of skewness also with webkb-flattened dataset (Figure \ref{fig-webkb-flattened-keywords}).
The skewness means that there is an imbalance of categories, which also influences the imbalance of topics.
However, in case of webkb-flattened dataset (Figure \ref{fig-webkb-flattened-keywords}), the number of categories is smaller than in Reuters-21578 dataset.
Thus, we can compensate the topical imbalance by extracting more keywords, as we can observe in Figure \ref{fig-webkb-flattened-keywords} by looking at the macro-averaged F1 score.

\begin{figure}[htb]
	\centering
	\begin{subfigure}[htb]{0.49\textwidth}
		\includegraphics[width=0.95\textwidth]{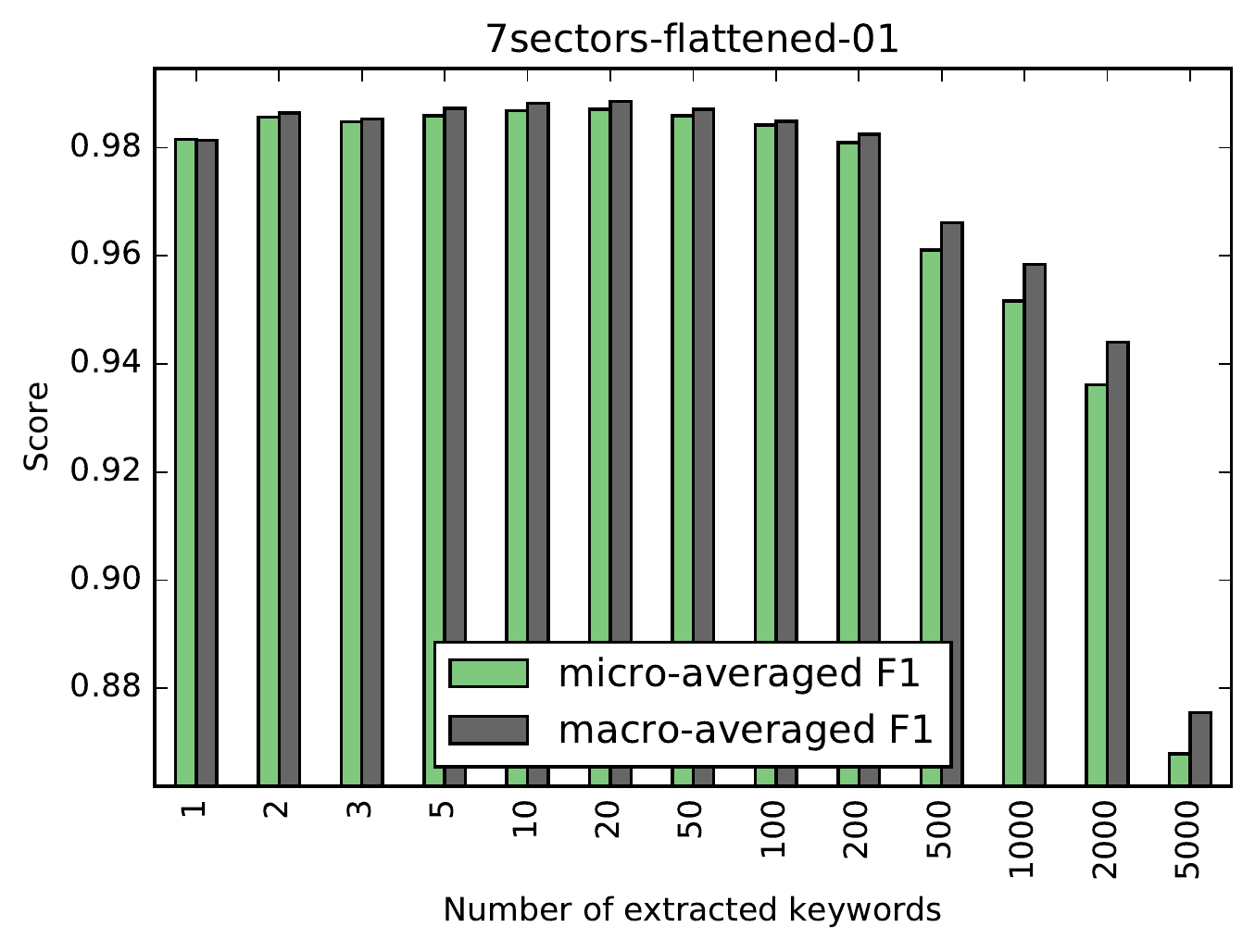}
	\end{subfigure}
	\begin{subfigure}[htb]{0.49\textwidth}
		\includegraphics[width=0.95\textwidth]{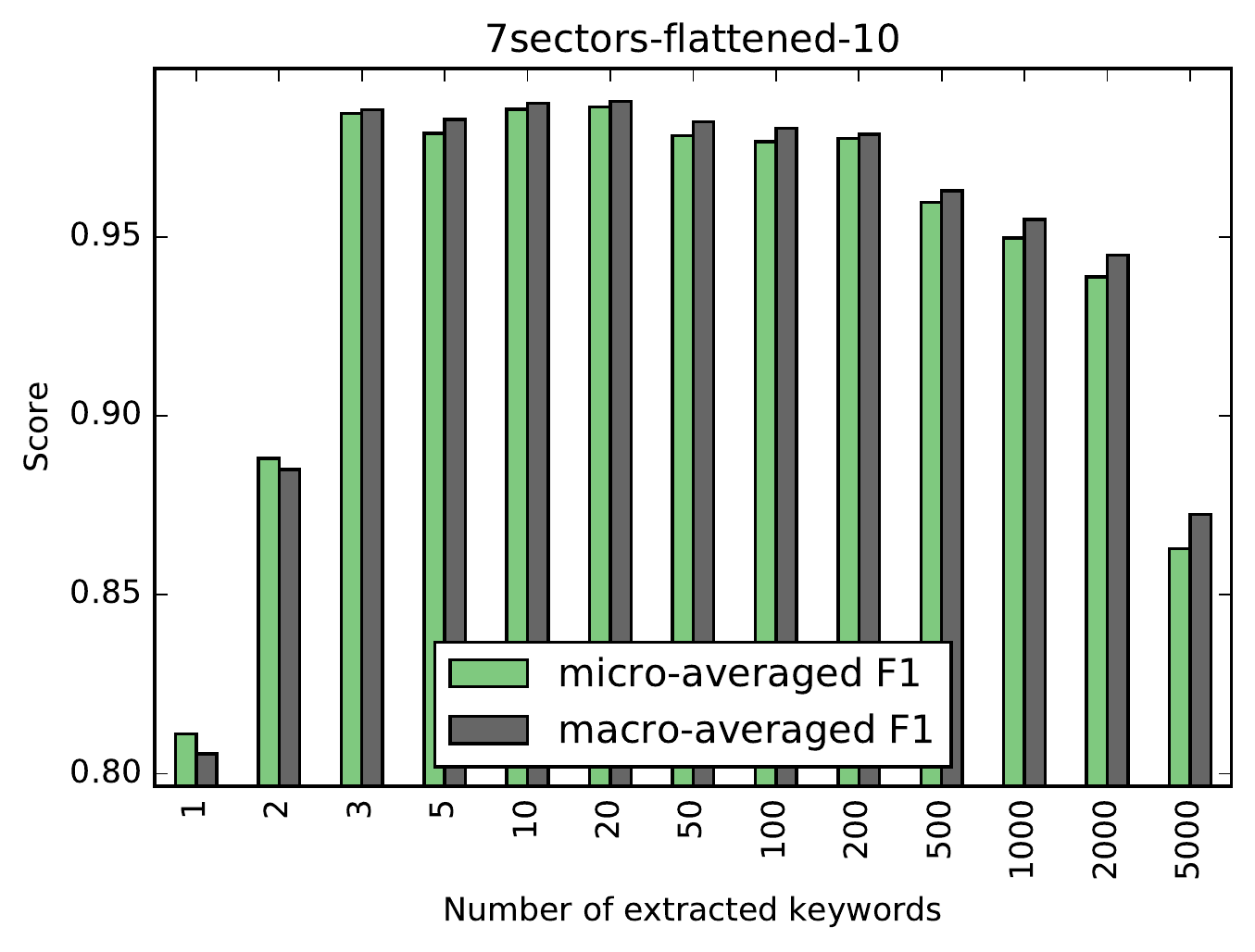}
	\end{subfigure}
	\caption{Comparison of proposed method's results for different number of extracted keywords evaluated on 7sectors-flattened dataset.}
	\label{fig-7sectors-flattened-keywords}
\end{figure}

With 7sectors dataset (\ref{fig-7sectors-flattened-keywords}), we can see how a low number of balanced categories enables high F1 score (both micro-averaged and macro-averaged) even with only one extracted keyword.

\subsection{Examples of extracted keywords}
Here, we present some examples of extracted keywords for our methods with different word-weighting metrics.
We focus on first two datasets 20newsgroups and Reuters-21578, which have more categories and give us a representative notion of how the extracted keywords look like.
From 20-newsgroups dataset, we choose the first document from \textit{comp.graphics} category as an example document.
It contains a message about video playback performance for different resolutions (see \ref{appendix-sample-documents}).
In Table \ref{table-top10-20newsgroups-ABCD}, we present extracted keywords using several different word-weighting metrics based on $ABCD$-statistics (discussed in Section \ref{sec-related-work}).
While all these metrics result in reasonable keywords extracted by our method, we can observe that metrics enriched with term frequency factor (whose signature has the additional \textit{tf} prefix) capture the specific topic discussed in the document better (pixel, encoding, graphics, video) than the others without it.

\begin{table}[h]
	\caption{\label{table-top10-20newsgroups-ABCD} 10 keywords for first document in category $comp.graphics$ of 20-newsgroups dataset for different word-weighting metrics based on $ABCD$-statistics.}
	\begin{center}
		\small
		\begin{tabular}{lp{0.75\textwidth}}
			\toprule
			\bf Metric & \bf Keywords \\ \midrule
			\textit{tf-rf} & \textit{pixel, pixels, DVD, outputs, bandwidth, CPU, encoding, graphics, interface, video} \\ \midrule
			\textit{rf} & \textit{rendering, animation, polygon, volumetric, magnification, dimensional, visualization, color, vertex, gradients} \\ \midrule
			\textit{tf-tds} & \textit{pixel, pixels, encoding, graphics, vertex, outputs, DVD, camera, tonal, aperture} \\ \midrule
			\textit{tds} & \textit{visualization, rendering, polygon, animation, gradients, magnification, vertex, volumetric, imagery, textures} \\ \midrule
			\textit{tf-ig} & \textit{pixel, pixels, encoding, graphics, DVD, outputs, vertex, video, camera, aperture} \\ \midrule
			\textit{ig} & \textit{vertex, polygon, rendering, visualization, animation, pixels, vertices, pixel, magnification, textures} \\ \bottomrule
		\end{tabular}
	\end{center}
\end{table}

Other word-weighting metrics (see Table \ref{table-top10-20newsgroups-other}) that are not based on $ABCD$-statistics do not appear to be very satisfying.
We can see that the term frequency factor helps to improve the quality of extracted terms, which means that their discriminative power is not so substantial as it is in case of the word-weighting metrics showcased in Table \ref{table-top10-20newsgroups-ABCD}.
We also include the results of \textit{tf-idf}, which differs from all other metrics in its nature, since it does not utilise a discriminative statistics.
However, we can see that it also produces very reasonable and relevant keywords (\textit{pixel, encoding, graphics, frames, discs} are all relevant), which are also quite discriminative despite the fact that \textit{tf-idf} is not a discriminative metric (see the discussion in Section \ref{sec-related-work}).

\begin{table}[h]
	\caption{\label{table-top10-20newsgroups-other} 10 keywords for first document in category comp.graphics of 20-newsgroups dataset for different discriminative metrics.}
	\begin{center}
		\small
		\begin{tabular}{lp{0.75\textwidth}}
			\toprule
			\bf Metric & \bf Keywords \\ \midrule
			\textit{tf-tf-dcf} & \textit{undisturbed, rendering, vertex, unprepared, volumetric, theater, movable, underdeveloped, striking, nonverbal} \\ \midrule
			\textit{tf-dcf} & \textit{undisturbed, unprepared, haphazard, runaway, underdeveloped, widest, silent, picket, kinetic, movable} \\ \midrule
			\textit{tf-thd} & \textit{DVD, bandwidth, CPU, pixel, outputs, encoding, pixels, interface, graphics, browser} \\ \midrule
			\textit{thd} & \textit{prosecutors, starter, religious, player, pitch, myth, souls, pitcher, opponent, saint} \\ \midrule
			\textit{tf-idf} & \textit{pixel, pixels, DVD, outputs, bandwidth, encoding, CPU, graphics, frames, discs} \\ \bottomrule
		\end{tabular}
	\end{center}
\end{table}

From Reuters-21578 dataset, we choose the first document from ship category as an example document.
It contains a message reporting a dispute about a ban on foreign-flag ships in Australian ports (see \ref{appendix-sample-documents}).
In Table \ref{table-top10-Reuters-21578-ABCD}, we present extracted keywords using the same discriminative metrics based on $ABCD$-statistics as in Table \ref{table-top10-20newsgroups-ABCD}.
Again, we can observe that our method based on these metrics gives reasonable results.
We can also notice the similar effect of term frequency factor as in the previous case.
This time, however, only \textit{tf-rf} metric captures non-discriminative words that are quite relevant (Australian cities). On the other hand, those metrics that do not contain the term frequency factor, capture more hidden concepts that are not mentioned directly (e.g., pirates, which describes protesting workers in ports).

\begin{table}[h]
	\caption{\label{table-top10-Reuters-21578-ABCD} 10 keywords for first document in category ship of Reuters-21578 dataset for different discriminative metrics based on $ABCD$-statistics.}
	\begin{center}
		\small
		\begin{tabular}{lp{0.75\textwidth}}
			\toprule
			\bf Metric & \bf Keywords \\ \midrule
			\textit{tf-rf} & \textit{cargo, Queensland, Brisbane, cargoes, Adelaide, Australian, Canberra, Victorian, Sydney, Perth} \\ \midrule
			\textit{rf} & \textit{Ship, Ships, anchorage, destroyers, steamer, destroyer, coastline, Seas, pirate, pirates} \\ \midrule
			\textit{tf-tds} & \textit{anchorage, Ships, Ship, steamer, vessels, maritime, port, ships, ship, dock} \\ \midrule
			\textit{tds} & \textit{pirates, pirate, steamer, anchorage, Ship, sail, ferry, destroyer, Ships, destroyers} \\ \midrule
			\textit{tf-ig} & \textit{Ship, Ships, anchorage, dock, ferry, maritime, ships, vessels, docks, port} \\ \midrule
			\textit{ig} & \textit{anchorage, destroyers, pirate, ferry, destroyer, Ship, pirates, steamer, Ships, nos} \\ \bottomrule
		\end{tabular}
	\end{center}
\end{table}

Previously discussed characteristics of other discriminative metrics (discussed for Table \ref{table-top10-20newsgroups-other}) appear to hold likewise also with the Reuters-21578 dataset, where we can observe similarly noisy keywords (Table \ref{table-top10-Reuters-21578-other}). This time, only \textit{tf-thd} metric seems to be improved by term frequency factor.
We consider keywords extracted using \textit{tf-idf} metric even more relevant than those extracted by using other metrics listed in Table \ref{table-top10-Reuters-21578-other}.
Apart from Australian cities, there is also a very relevant word ports and one of the most relevant words, cargo, is higher in the list.

\begin{table}[h]
	\caption{\label{table-top10-Reuters-21578-other} 10 keywords for first document in category ship of Reuters-21578 dataset for different importance metrics.}
	\begin{center}
		\small
		\begin{tabular}{lp{0.75\textwidth}}
			\toprule
			\bf Metric & \bf Keywords \\ \midrule
			\textit{tf-tf-dcf} & \textit{nos, Battle, extant, ridges, metrical, locomotion, mouse, cheers, slippers, Nellie} \\ \midrule
			\textit{tf-dcf} & \textit{Battle, extant, destroyers, Pilot, leafy, ridges, wings, pirate, cheers, pirates} \\ \midrule
			\textit{tf-thd} & \textit{Brisbane, Canberra, Melbourne, Perth, Sydney, cargo, Australian, Adelaide, cargoes, Auckland} \\ \midrule
			\textit{thd} & \textit{rent, owned, finish, files, Paid, adjustments, proceeding, foregoing, operators, resumption} \\ \midrule
			\textit{tf-idf} & \textit{cargo, Queensland, Brisbane, Victorian, Perth, Adelaide, Sydney, ports, Canberra, cargoes} \\ \bottomrule
		\end{tabular}
	\end{center}
\end{table}

\subsection{Manual quantitative evaluation}
We also performed a manual quantitative evaluation of extracted keywords on the 20-newsgroups dataset.
We developed a custom experimental web application to gather assessments of extracted keywords from human judges.
The participation in the experiment was voluntary, and most of the judges that participated in the experiment had computer science background, which made them more suitable as the 20-newsgroups dataset contains multiple computer-related categories.
We randomly chose five articles from each category.
We decided to choose one representative from discriminative word-weighting metrics and one representative variant of our method.
For each article, we took 5 keywords extracted by \textit{tf-rf} metric and 5 keywords extracted by our \textit{latent-5word-knn90-rf} method.
We presented the keywords as a one set, so judges did not know which method extracted which keyword.

To gather enough assessments in a controlled manner, we split categories into batches (5 categories per batch) and three different judges assessed each document.
We designed the methodology in such a way so that we would gather as complete assessments of documents within batches as possible.
We fixed the order of batches and wanted to gather all assessments for documents in the first batch before advancing to the next one.
For each judge, we presented the documents from the current batch in random order.
If a single judge assessed all documents in the current batch or there were no more document in the current batch with less than three assessments, the next batch for the user was opened, and she could assess documents from that next batch.

In the experiment, we started off by introducing the instructions to each judge to clarify the objective.
Then we presented each judge with article text, all categories in the dataset with the categories that the article belongs to highlighted in green and unique keywords extracted by the two methods shuffled in random order.
We asked each judge two questions:

\begin{enumerate}
	\item Can it decide article topic?
	\item Can it decide article category?
\end{enumerate}

Then we required an answer for each keyword to both questions.
We accomplished that by disabling submit button until all answers were filled in.
The purpose of this is to avoid skewed results by pre-filling the answers if the judge missed out an answer.
The possible answers to assess the quality of an extracted keyword were \textit{Do not know, Irrelevant, Partially} and \textit{Relevant}.

\begin{figure}[htb]
	\centering
	\resizebox{\textwidth}{!}{%
		\includegraphics[width=7cm]{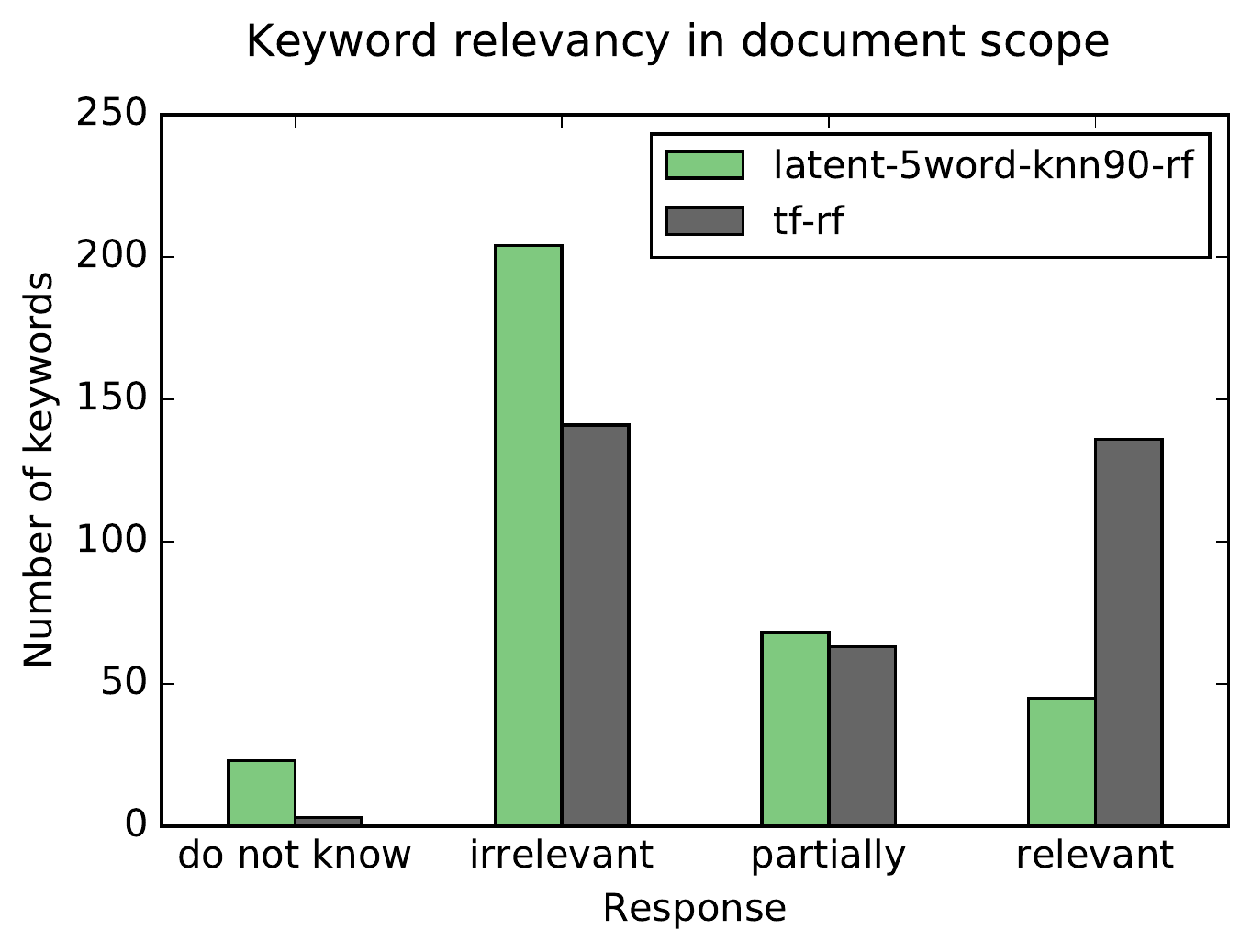}
		\includegraphics[width=7cm]{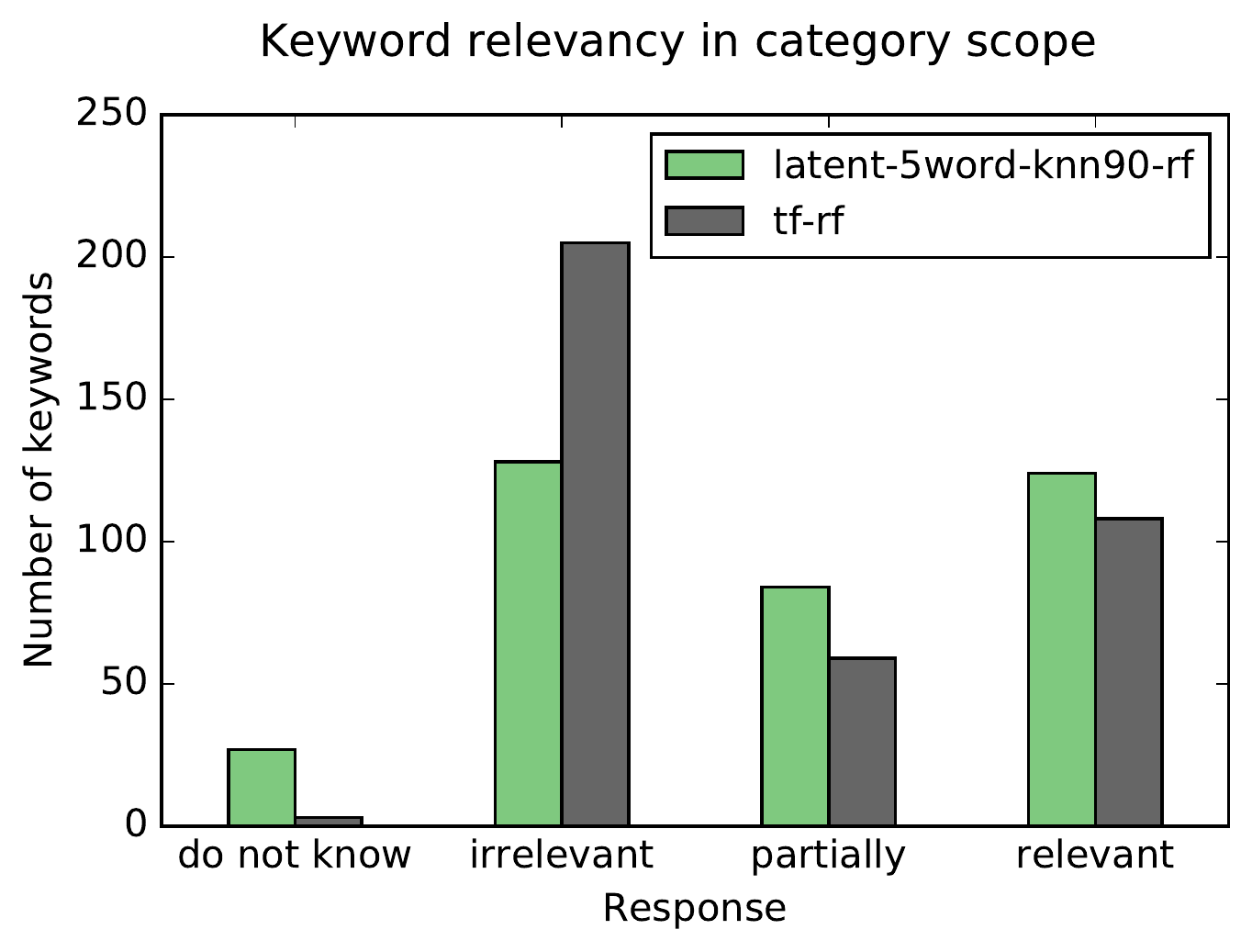}
	}
	\caption{Human judges ratings - comparing our \textit{latent-5word-knn90-rf} method and \textit{tf-rf}.}
	\label{fig-human-ratings}
\end{figure}

\begin{figure}[htb]
	\centering
	\includegraphics[width=6cm]{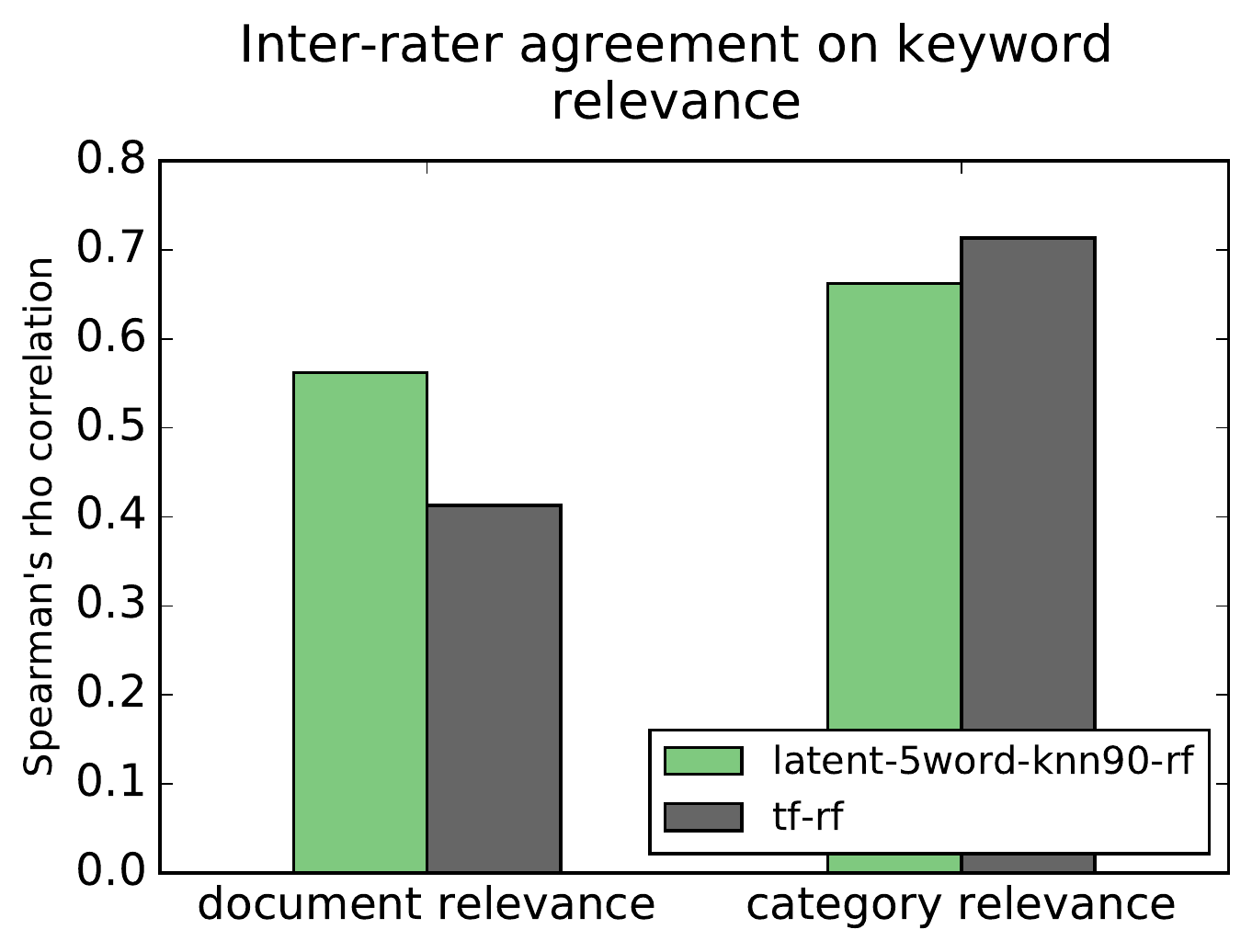}\\
	\caption{Correlation of human judges ratings - comparing our \textit{latent-5word-knn90-rf} method and \textit{tf-rf}.}
	\label{fig-inter-rater_agreement}
\end{figure}

In Figure \ref{fig-human-ratings} we compare answers of human judges to both questions assessing keywords produced by methods \textit{latent-5word-knn90-rf} and \textit{tf-rf}.
We chose \textit{tf-rf} as a baseline method since it was reported to give best results on 20newsgroups dataset among the frequency-based methods which we build on \cite{Lan:2006}.
In Figure \ref{fig-inter-rater_agreement} we compare correlation of human ratings for our \textit{latent-5word-knn90-rf} method and \textit{tf-rf} method.
We evaluate the inter-rater agreement by Spearman's correlation coefficient \citep{Kokoska:2000:crc} since we deal with ordinal responses and can omit the response by giving \textit{do not know} response.

In Figure \ref{fig-human-ratings} we can see that our method is biased to be more discriminative, which results in being more related to the category, but not so relevant to the exact document.
Looking at the inter-rater agreement in Figure \ref{fig-inter-rater_agreement}, we can see that our statement is also validated by the correlation with human responses.
Although the baseline \textit{tf-rf} metric has more relevant keywords according to Figure \ref{fig-human-ratings}, we can see that it does not correlate well with human responses on document relevance (Figure \ref{fig-inter-rater_agreement}), suggesting that the document relevance of those keywords is disputable and weak.
In Table \ref{table-human-ratings-significance}, we provide the results of statistical significance tests.
As we can see, all the results are statistically highly significant with the p-value less than 0.001.

\begin{table}[htb]
	\caption{Statistical significance of human ratings.}
	\label{table-human-ratings-significance}
	\begin{center}
		\footnotesize
		\begin{tabular}{llrr}
			\toprule
			\multicolumn{2}{c}{\bf Significance test} & \bf Document relevance & \bf Category relevance \\
			\midrule
			\multirow{3}{*}{\begin{minipage}{0.8in}Wilcoxon signed-rank test\end{minipage}} & T statistics & 1299847.5 & 1871991 \\
				\cmidrule{2-4}
			& z statistics & -17.37 & -3.53 \\
			\cmidrule{2-4}
			& two-sided p-value & 1.36e-67 & 4.2e-4 \\
			\midrule
			\multirow{2}{*}{Sign-test} & M statistics & -438.5 & 97.5 \\
			\cmidrule{2-4}
			& one-sided p-value & 7.02e-62 & 2.7e-4 \\
			\bottomrule
		\end{tabular}
	\end{center}
\end{table}

\section{Conclusion}
We have presented a novel method for keyword extraction, which takes advantage of machine-learned latent features of words as a source of semantics.
We have shown that our method can be used to extract a small number of keywords to obtain a human-readable form of metadata that summarises the whole article.
However, operating in multidimensional latent feature space, we can easily scale up to improve the performance of automated processing of documents, which can handle a much higher number of keywords by the simple transformation of keywords into one document feature vector.
We consider such scalability of our keyword extraction as a significant contribution since up to now, it was not possible with other pure word-weighting methods.

On top of the concise representation, we have demonstrated that our method is capable of categorising a document with high precision, which proves that such representation is also highly discriminative between various topics.
Our method can utilise various discriminative metrics and improve their performance while using compact representation.
The main limitation of the proposed method is that we require the existence of categories which can be used to categorise documents.
The most notable contribution is state-of-the-art results, e.g. on the 20-newsgroups dataset, we achieve $99.32$\% micro-F1 score using such concise representation as just $10$ keywords per document.

Our method has an extensive range of applications, as it turns out that many fields of interests are naturally subdivided into multiple categories.
For example, most web fora and newspaper sites contain multiple topical categories, social networks contain multiple social groups, and archives of research papers are divided into several research fields.
However, even if we do not have any labels for sub-categories (e.g., because they are latent), thanks to the chosen distributed representation, we can easily compute keywords for any set of documents.
Like that, we can compute relevant keywords for a category as a whole, similarly like latent topic models (like latent Dirichlet allocation) do.
In particular, it seems like an interesting idea to combine our method with latent topic models.
Our method could help in making the topics more discriminative, which could enhance the quality of probability distributions over words within the individual topics.

Our keyword extraction method is highly flexible, which means that we can fine-tune it for specific needs and data, but also that one must be careful to set up the parameters.
We have shown that chosen discriminative metric can be rather influential.
Looking ahead, it would be interesting to experiment with means of how to automatically find or learn an optimal combination of four statistical variables $ABCD$-discussed in section \ref{sec-related-work}.
We would also like to experiment with utilising the discriminative metrics earlier in the process to obtain more category-dependent similar words for given candidate phrase.

{\bf Acknowledgement.} This work was partially supported by the Slovak Research and Development Agency under the contract No. APVV-15-0508, the Scientific Grant Agency of the Slovak Republic, grant No. VG 1/0646/15 and the Research and Development Operational Programme, ITMS 26240120039, co-funded by the ERDF.

\section*{\refname}
\bibliographystyle{elsarticle-harv}
\bibliography{main.bib}

\appendix

\clearpage

\section{Example phrases extracted by our automaton}
\label{appendix-pospatterns-phrases}

Here, we present examples of extracted phrases by the automaton in Figure \ref{fig-phrase-automaton}.
For each pattern of extracted POS tags, we first chose ten random phrases.
Then we randomly selected 50 phrases of increasing length.
Note that the proposed Algorithm \ref{algo-phrase_vector} skips all words that are not present in the model of latent feature vectors.

\begin{table}[htb]
	\caption{Extracted phrases and the respective POS tag patterns.}
	\tiny
	\begin{tabular}{ll}
		\toprule
		\bf POS tag pattern & \bf extracted phrase \\
		\midrule
JN & Mathematical Foundations \\
NN & Searching Tools \\
JIN & better than competition \\
JJN & Macedonian human rights \\
JNN & small Ottoman garrison \\
NIN & Design of Pyramid \\
NJN & toll free help \\
JIJN & highest among young males \\
JINN & > In article <1993May10.162032.3955@colorado.edu> \\
JJJN & separate real-time wireframe viewer \\
JJNN & fast state-of-the-art factorization algorithms \\
NINN & Factors in Computing Systems \\
NJJN & eleven international scientific societies \\
JIJIN & > aab > with X11 \\
JIJJN & responsible for salivary inhibitory activity \\
JIJNN & true for other encryption devices \\
JININ & overwhelming at times with posts \\
JINNN & traditional under water lighting products \\
JJIJN & MORE effective in similar cases \\
JJINN & > > In article <93108.025818U28037@uicvm.uic.edu> \\
JJJIN & systemic anti-fungal such as itraconazole \\
JJJJN & several other lesser known cults \\
JJJNN & cumulative redeemable convertible series D \\
JJNIN & international ecumenical community of monks \\
JJNJN & Greek nationalist movement Philike Etairia \\
JJNNN & largest independent farm service center \\
JNIJN & good debuggers for parallel systems \\
JNINN & bitter conflict between Sri Lanka \\
JNJIN & Full version available as Cornell \\
JNJJN & intermediate level human visual sensors \\
JNNIN & templated data types in STL \\
JNNJN & original IBM foam fitted boxies \\
JNNNN & Energetic Gamma Ray Experiment Telescope \\
NIJIN & absence of good as evil \\
NIJJN & efforts in advanced technical ceramics \\
NIJNN & information with high speed manipulator \\
NININ & Center for Excellence in Space \\
NINJN & applications of fiber optic lighting \\
NINNN & Lu at Sun Sun Nov \\
NJIJN & rights such as collective bargaining \\
NJJIN & fit routine useful for interpolation \\
NJJJN & debt heavy Latin American producer \\
NJJNN & bank net foreign currency assets \\
NJNIN & PC central display of Respitrace \\
NNIJN & System Edits of Eligible Versus \\
NNINN & International Journal of Robotics Research \\
NNJIN & container classes such as lists \\
NNJJN & Orientation sessions last 60-90 minutes \\
NNNIN & Austin \#\# Department of Computer \\
NNNNN & PCT ORANGE-CO <OJ> STAKE WASHINGTON \\
\bottomrule
	\end{tabular}
\end{table}

\clearpage

\section{Multiword phrases in the feature vector model}
\label{appendix-w2vphrases}

Here, we provide examples of multiword phrases in the used feature vector model\footnote{
Available at \url{https://code.google.com/archive/p/word2vec/} in February 22, 2017
}.
Out of 3 million terms, there are $2\,070\,978$ multiword phrases, i.e. more than two-thirds.
For each multiword phrase, we compare the differences between the nearest neighbours of the phrase vector and the sum of the individual word vectors.

	\begin{center}
		\small
		\begin{longtable}{p{0.15\textwidth}p{0.35\textwidth}p{0.35\textwidth}}
	\caption{
		Examples of multiword phrases in used feature vector model.
	}
	\label{table-w2vphrases} \\
			\toprule
			\bf Multiword phrase & \bf 10 nearest neighbours for the phrase vector & \bf 10 nearest neighbours for the sum of the individual word vectors \\ \midrule

\textit{Tom Cruise} &
\textit{Angelina Jolie, Brad Pitt, Cruise, David Beckham, Lindsay Lohan, Jolie, Britney Spears, Madonna, movie, Hollywood} &
\textit{Bob, Jim, Rob, Mike, Tim, Steve, Greg, Ken, Brian, Dave} \\ \midrule

\textit{Wall Street} &
\textit{Dow Jones industrial, Bear Stearns, Dow Jones Industrial Average, Goldman Sachs, stocks, investors, earnings, Stocks, analysts, Lehman Brothers} &
\textit{Streets, Avenue, Main Street, Main, St., street, Square, Ave, streets, Road} \\ \midrule

\textit{Middle East} &
\textit{Mideast, Middle Eastern, Asia, Arab, Africa, Latin America, Southeast Asia, Saudi Arabia, Europe, Persian Gulf} &
\textit{South, West, North, Central, Upper, Northeast, Southeast, Eastern, Lower, middle} \\ \midrule

\textit{United States} &
\textit{U.S., America, Europe, countries, Canada, United Kingdom, country, world, Great Britain, Latin America} &
\textit{states, Inter, Countries, Union, Agencies, Governments, nations, Continental, Arsenal, Everton} \\ \midrule

\textit{Social Security} &
\textit{Medicare, pensions, retirees, pension, Medicaid, health care, retiree, paycheck, AARP, welfare} &
\textit{security, Employment, Intelligence, Protection, Integration, Communication, Health, Economic, social, Criminal} \\ \midrule

\textit{hip hop} &
\textit{rap, rapper, jazz, punk, music, Jay Z, rock n roll, indie, Eminem, gospel} &
\textit{knee, ankle, leg, shoulder, heel, thigh, elbow, wrist, groin, toe} \\ \midrule

\textit{semi final} &
\textit{semi finals, finals, semifinal, quarterfinal, quarterfinals, semifinals, semis, qualifier, match, Finals} &
\textit{Final, second, first, third, fourth, finals, sixth, seventh, fifth, eighth} \\ \midrule

\textit{New Year Eve} &
\textit{Christmas Eve, Valentine Day, Halloween, Christmas, holiday, Thanksgiving, birthday, night, festive, midnight} &
\textit{Day, Happy, Week, Deal, Christmas, Years, Thanksgiving, Month, holiday, Best} \\ \midrule

\textit{red carpet} &
\textit{backstage, gala, Oscars, Academy Awards, gown, Hollywood, celebrities, runway, onstage, dress} &
\textit{blue, yellow, purple, pink, brown, colored, green, orange, gray, white} \\ \midrule

\textit{International Space Station} &
\textit{astronauts, ISS, spacecraft, space shuttle, NASA, Endeavour, astronaut, orbit, shuttle, Atlantis} &
\textit{station, National, Global, Aviation, Maritime, Technology, Facility, Headquarters, Logistics, World} \\ \bottomrule

		\end{longtable}
	\end{center}

\clearpage

\section{Distribution of documents over categories}
\label{appendix-datasets}


\begin{figure}[htb]
	\centering
	\includegraphics[width=\textwidth]{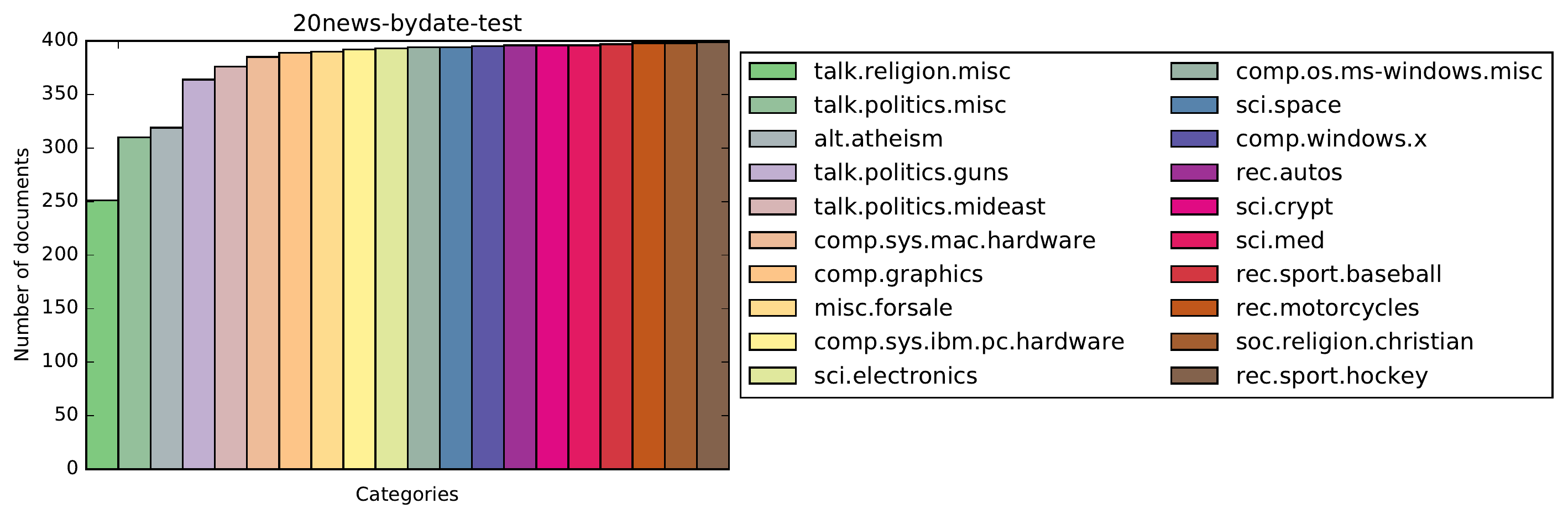}
%
	\includegraphics[width=\textwidth]{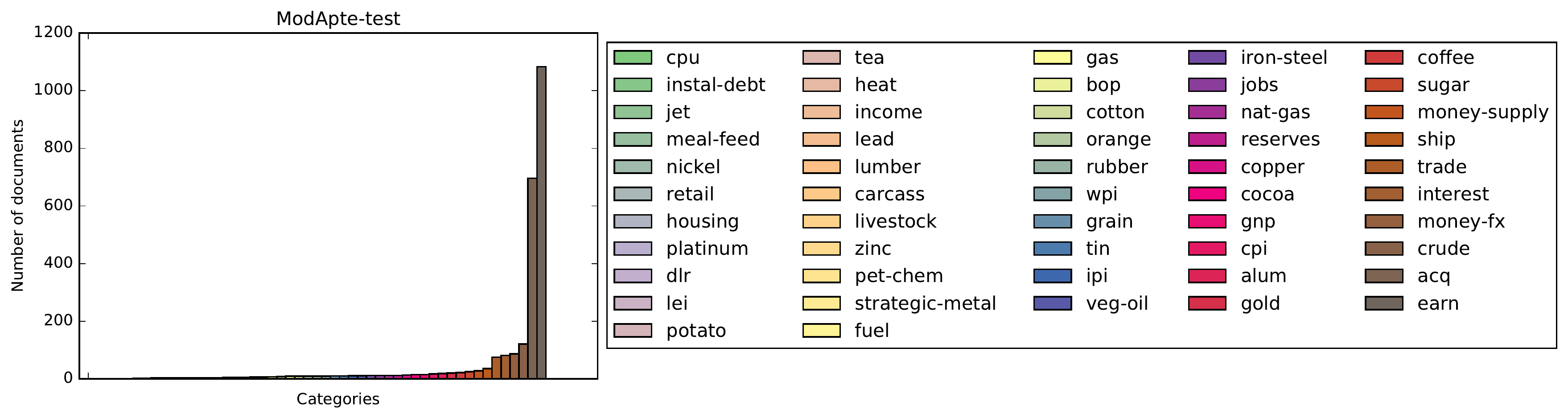}
%
	\includegraphics[width=0.45\textwidth]{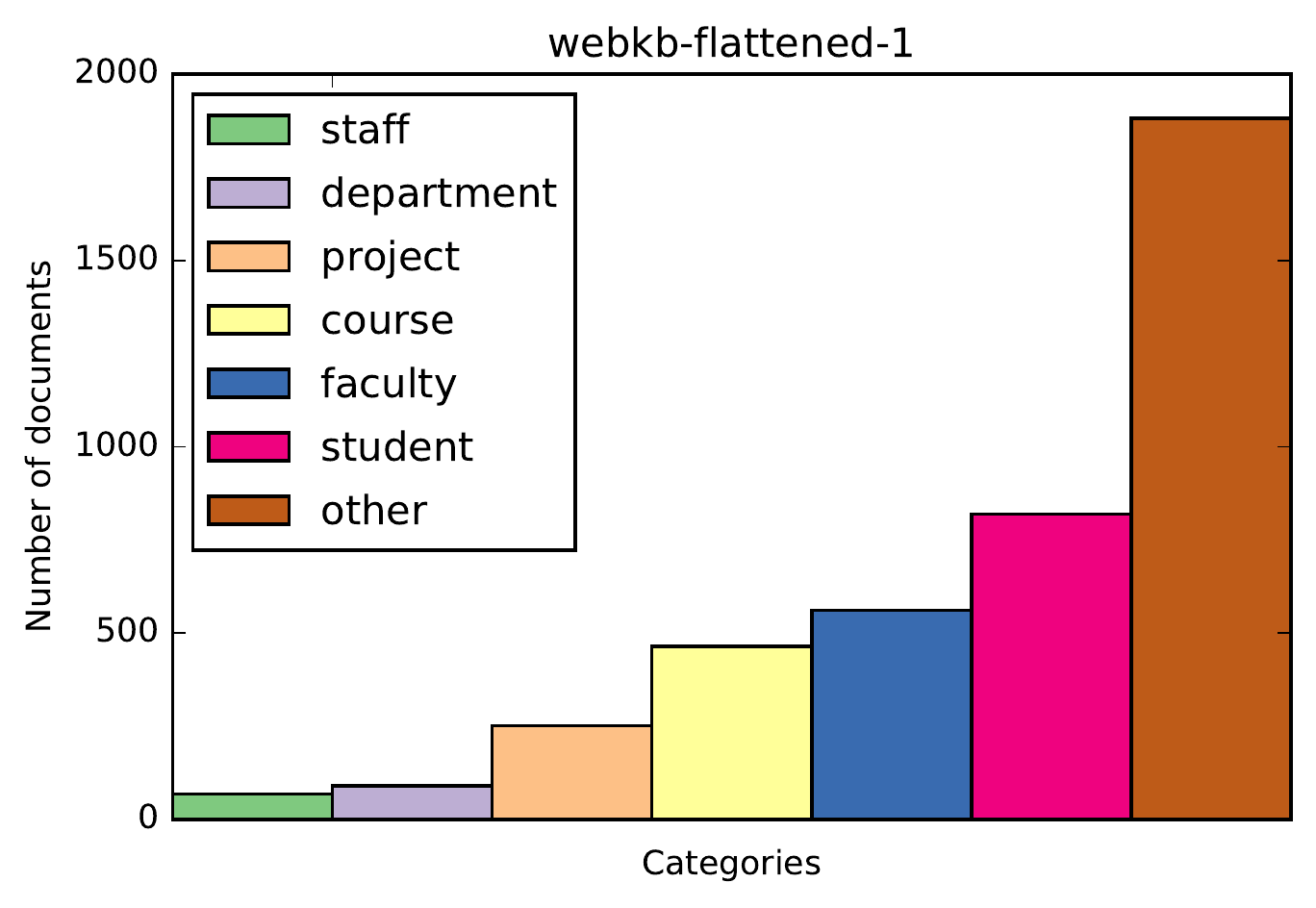}
	\includegraphics[width=0.45\textwidth]{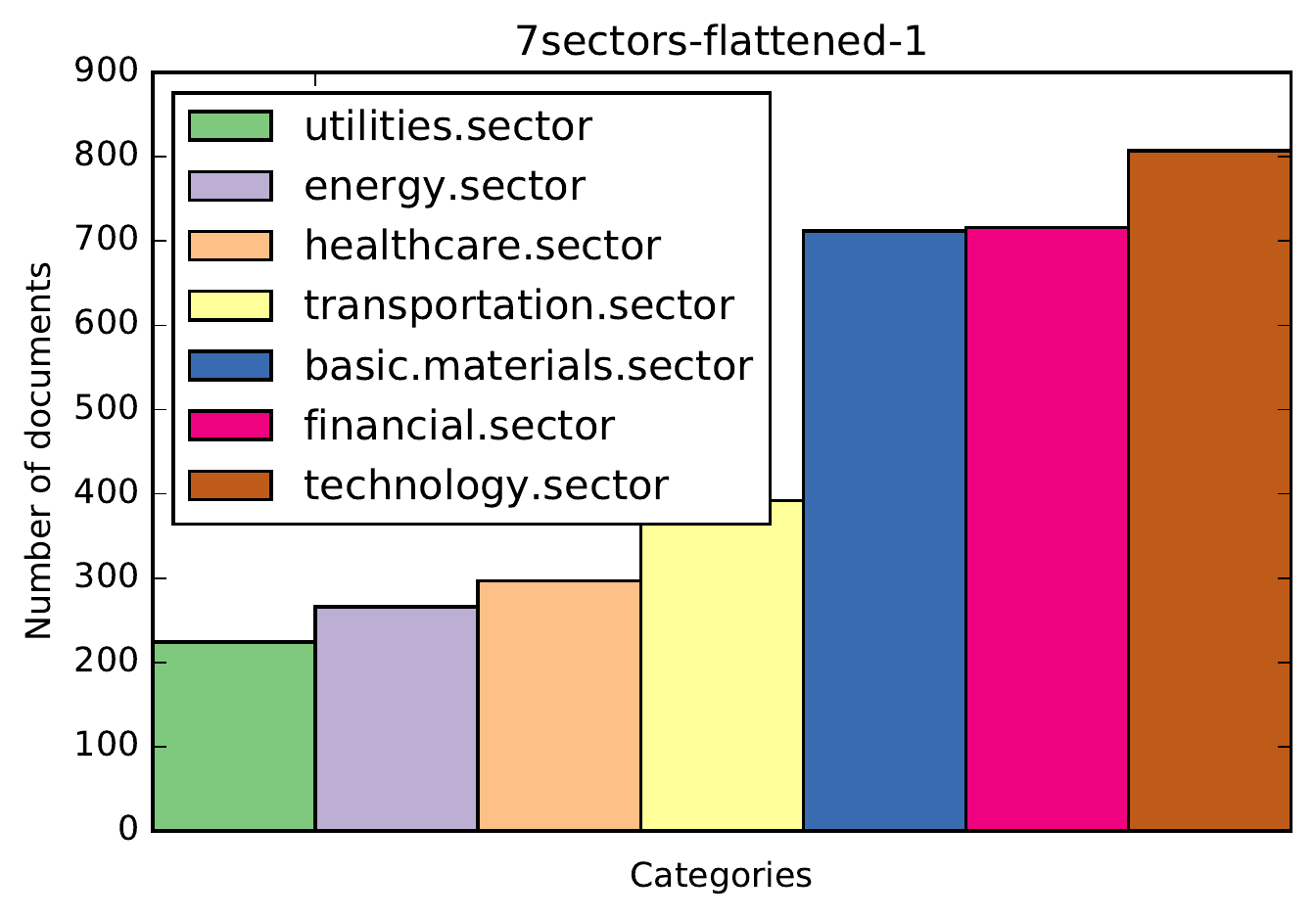}
	\label{fig-skewness-datasets}
	\caption{Distribution of documents over categories in evaluated datasets. For every dataset, the distribution is almost identical between the training and testing set.}
\end{figure}


\clearpage

\section{Comparing statistical significance for different metrics used in word ranking}
\label{appendix-metrics-significance}

Here, we compare statistical significance for different metrics used in word ranking.
First, we perform Wilcoxon signed-rank test \citep{Kokoska:2000:crc}.
In Table \ref{table-metrics-significance}, we report T statistics, z statistics and two-sided p-value for each pair of word-weighting metrics.
We supplement this with one-tailed sign test in Table \ref{table-metrics-significance2}, where we report M statistics and one-sided p-value.
In all cases, the sample size $N = 24$, we apply a Bonferroni correction to testing multiple comparisons, and we highlight statistically significant differences in bold.

\afterpage{%
	\clearpage
	\thispagestyle{empty}
		\begin{landscape}

\begin{table}[htb]
	\tiny
	\caption{
		Statistical significance of micro-averaged F1 score differences for pairs of word-weighting metrics when used in our method using two-tailed Wilcoxon signed-rank test.
	}
	\label{table-metrics-significance}
	\tiny
	\begin{tabular}{lcccccccccccccc}
		\toprule
		{} & \textit{tf-idf} & \textit{tf-ig} & \textit{tf-rf} & \textit{tf-tds} & \textit{tf-tf-dcf} & \textit{tf-thd} & \textit{tf-dcf} & $ig$ & $rf$ & $tds$ & $thd$ & $tf\mhyphen \chi^2$ & $\chi^2$ \\
\midrule
\midrule
		\textit{tf-idf} &                                - &      \begin{tabular}{@{}c@{}} \bf 0.0\\ \bf -4.29\\ \bf 2.84e-3\end{tabular} &      \begin{tabular}{@{}c@{}} \bf 5.0\\ \bf -4.14\\ \bf 5.35e-3\end{tabular} &      \begin{tabular}{@{}c@{}} \bf 0.0\\ \bf -4.29\\ \bf 2.84e-3\end{tabular} &      \begin{tabular}{@{}c@{}} \bf 4.0\\ \bf -4.17\\ \bf 4.72e-3\end{tabular} &     \begin{tabular}{@{}c@{}} \bf 12.0\\ \bf -3.94\\ \bf 1.26e-2\end{tabular} &      \begin{tabular}{@{}c@{}} \bf 7.0\\ \bf -4.09\\ \bf 6.85e-3\end{tabular} &     \begin{tabular}{@{}c@{}} \bf 0.0\\ \bf -4.29\\ \bf 2.84e-3\end{tabular} &     \begin{tabular}{@{}c@{}} \bf 0.0\\ \bf -4.29\\ \bf 2.84e-3\end{tabular} &     \begin{tabular}{@{}c@{}} \bf 0.0\\ \bf -4.29\\ \bf 2.84e-3\end{tabular} &      \begin{tabular}{@{}c@{}} \bf 8.0\\ \bf -4.06\\ \bf 7.75e-3\end{tabular} &   \begin{tabular}{@{}c@{}} 76.0\\ -2.11\\ 1.00e+00\end{tabular} &  \begin{tabular}{@{}c@{}} 146.0\\ -0.11\\ 1.00e+00\end{tabular} \\
\midrule
		\textit{tf-ig} &      \begin{tabular}{@{}c@{}} \bf 0.0\\ \bf -4.29\\ \bf 2.84e-3\end{tabular} &                                - &     \begin{tabular}{@{}c@{}} \bf 12.0\\ \bf -3.94\\ \bf 1.26e-2\end{tabular} &  \begin{tabular}{@{}c@{}} 135.5\\ -0.41\\ 1.00e+00\end{tabular} &   \begin{tabular}{@{}c@{}} 56.0\\ -2.49\\ 1.00e+00\end{tabular} &      \begin{tabular}{@{}c@{}} \bf 0.0\\ \bf -4.29\\ \bf 2.84e-3\end{tabular} &   \begin{tabular}{@{}c@{}} 80.0\\ -2.00\\ 1.00e+00\end{tabular} &     \begin{tabular}{@{}c@{}} \bf 0.0\\ \bf -4.20\\ \bf 4.21e-3\end{tabular} &     \begin{tabular}{@{}c@{}} \bf 0.0\\ \bf -4.29\\ \bf 2.84e-3\end{tabular} &     \begin{tabular}{@{}c@{}} \bf 0.0\\ \bf -4.29\\ \bf 2.84e-3\end{tabular} &   \begin{tabular}{@{}c@{}} 87.0\\ -1.80\\ 1.00e+00\end{tabular} &   \begin{tabular}{@{}c@{}} 83.0\\ -1.91\\ 1.00e+00\end{tabular} &   \begin{tabular}{@{}c@{}} 63.0\\ -2.49\\ 1.00e+00\end{tabular} \\
\midrule
		\textit{tf-rf} &      \begin{tabular}{@{}c@{}} \bf 5.0\\ \bf -4.14\\ \bf 5.35e-3\end{tabular} &     \begin{tabular}{@{}c@{}} \bf 12.0\\ \bf -3.94\\ \bf 1.26e-2\end{tabular} &                                - &     \begin{tabular}{@{}c@{}} \bf 21.0\\ \bf -3.69\\ \bf 3.56e-2\end{tabular} &  \begin{tabular}{@{}c@{}} 110.0\\ -1.14\\ 1.00e+00\end{tabular} &      \begin{tabular}{@{}c@{}} \bf 0.0\\ \bf -4.29\\ \bf 2.84e-3\end{tabular} &     \begin{tabular}{@{}c@{}} \bf 23.0\\ \bf -3.63\\ \bf 4.45e-2\end{tabular} &     \begin{tabular}{@{}c@{}} \bf 0.0\\ \bf -4.29\\ \bf 2.84e-3\end{tabular} &     \begin{tabular}{@{}c@{}} \bf 0.0\\ \bf -4.29\\ \bf 2.84e-3\end{tabular} &     \begin{tabular}{@{}c@{}} \bf 0.0\\ \bf -4.29\\ \bf 2.84e-3\end{tabular} &    \begin{tabular}{@{}c@{}} 25.0\\ -3.57\\ 5.54e-2\end{tabular} &  \begin{tabular}{@{}c@{}} 136.0\\ -0.40\\ 1.00e+00\end{tabular} &  \begin{tabular}{@{}c@{}} 102.0\\ -1.37\\ 1.00e+00\end{tabular} \\
\midrule
		\textit{tf-tds} &      \begin{tabular}{@{}c@{}} \bf 0.0\\ \bf -4.29\\ \bf 2.84e-3\end{tabular} &  \begin{tabular}{@{}c@{}} 135.5\\ -0.41\\ 1.00e+00\end{tabular} &     \begin{tabular}{@{}c@{}} \bf 21.0\\ \bf -3.69\\ \bf 3.56e-2\end{tabular} &                                - &   \begin{tabular}{@{}c@{}} 55.0\\ -2.71\\ 1.00e+00\end{tabular} &      \begin{tabular}{@{}c@{}} \bf 0.0\\ \bf -4.29\\ \bf 2.84e-3\end{tabular} &   \begin{tabular}{@{}c@{}} 94.0\\ -1.60\\ 1.00e+00\end{tabular} &     \begin{tabular}{@{}c@{}} \bf 0.0\\ \bf -4.29\\ \bf 2.84e-3\end{tabular} &     \begin{tabular}{@{}c@{}} \bf 0.0\\ \bf -4.29\\ \bf 2.84e-3\end{tabular} &     \begin{tabular}{@{}c@{}} \bf 0.0\\ \bf -4.29\\ \bf 2.84e-3\end{tabular} &   \begin{tabular}{@{}c@{}} 89.0\\ -1.74\\ 1.00e+00\end{tabular} &   \begin{tabular}{@{}c@{}} 72.0\\ -2.23\\ 1.00e+00\end{tabular} &   \begin{tabular}{@{}c@{}} 64.0\\ -2.46\\ 1.00e+00\end{tabular} \\
\midrule
		\textit{tf-tf-dcf} &      \begin{tabular}{@{}c@{}} \bf 4.0\\ \bf -4.17\\ \bf 4.72e-3\end{tabular} &   \begin{tabular}{@{}c@{}} 56.0\\ -2.49\\ 1.00e+00\end{tabular} &  \begin{tabular}{@{}c@{}} 110.0\\ -1.14\\ 1.00e+00\end{tabular} &   \begin{tabular}{@{}c@{}} 55.0\\ -2.71\\ 1.00e+00\end{tabular} &                                - &      \begin{tabular}{@{}c@{}} \bf 1.0\\ \bf -4.26\\ \bf 3.23e-3\end{tabular} &    \begin{tabular}{@{}c@{}} 31.0\\ -3.40\\ 1.05e-1\end{tabular} &     \begin{tabular}{@{}c@{}} \bf 0.0\\ \bf -4.29\\ \bf 2.84e-3\end{tabular} &     \begin{tabular}{@{}c@{}} \bf 0.0\\ \bf -4.29\\ \bf 2.84e-3\end{tabular} &     \begin{tabular}{@{}c@{}} \bf 0.0\\ \bf -4.29\\ \bf 2.84e-3\end{tabular} &    \begin{tabular}{@{}c@{}} 46.0\\ -2.97\\ 4.62e-1\end{tabular} &  \begin{tabular}{@{}c@{}} 109.0\\ -1.17\\ 1.00e+00\end{tabular} &   \begin{tabular}{@{}c@{}} 77.0\\ -2.09\\ 1.00e+00\end{tabular} \\
\midrule
		\textit{tf-thd} &     \begin{tabular}{@{}c@{}} \bf 12.0\\ \bf -3.94\\ \bf 1.26e-2\end{tabular} &      \begin{tabular}{@{}c@{}} \bf 0.0\\ \bf -4.29\\ \bf 2.84e-3\end{tabular} &      \begin{tabular}{@{}c@{}} \bf 0.0\\ \bf -4.29\\ \bf 2.84e-3\end{tabular} &      \begin{tabular}{@{}c@{}} \bf 0.0\\ \bf -4.29\\ \bf 2.84e-3\end{tabular} &      \begin{tabular}{@{}c@{}} \bf 1.0\\ \bf -4.26\\ \bf 3.23e-3\end{tabular} &                                - &      \begin{tabular}{@{}c@{}} \bf 4.0\\ \bf -4.17\\ \bf 4.72e-3\end{tabular} &     \begin{tabular}{@{}c@{}} \bf 0.0\\ \bf -4.29\\ \bf 2.84e-3\end{tabular} &     \begin{tabular}{@{}c@{}} \bf 0.0\\ \bf -4.29\\ \bf 2.84e-3\end{tabular} &     \begin{tabular}{@{}c@{}} \bf 0.0\\ \bf -4.29\\ \bf 2.84e-3\end{tabular} &      \begin{tabular}{@{}c@{}} \bf 3.0\\ \bf -4.20\\ \bf 4.16e-3\end{tabular} &   \begin{tabular}{@{}c@{}} 65.0\\ -2.43\\ 1.00e+00\end{tabular} &  \begin{tabular}{@{}c@{}} 117.0\\ -0.94\\ 1.00e+00\end{tabular} \\
\midrule
		\textit{tf-dcf} &      \begin{tabular}{@{}c@{}} \bf 7.0\\ \bf -4.09\\ \bf 6.85e-3\end{tabular} &   \begin{tabular}{@{}c@{}} 80.0\\ -2.00\\ 1.00e+00\end{tabular} &     \begin{tabular}{@{}c@{}} \bf 23.0\\ \bf -3.63\\ \bf 4.45e-2\end{tabular} &   \begin{tabular}{@{}c@{}} 94.0\\ -1.60\\ 1.00e+00\end{tabular} &    \begin{tabular}{@{}c@{}} 31.0\\ -3.40\\ 1.05e-1\end{tabular} &      \begin{tabular}{@{}c@{}} \bf 4.0\\ \bf -4.17\\ \bf 4.72e-3\end{tabular} &                                - &     \begin{tabular}{@{}c@{}} \bf 8.0\\ \bf -3.95\\ \bf 1.20e-2\end{tabular} &   \begin{tabular}{@{}c@{}} 20.0\\ -3.59\\ 5.18e-2\end{tabular} &    \begin{tabular}{@{}c@{}} \bf 21.5\\ \bf -3.67\\ \bf 3.76e-2\end{tabular} &  \begin{tabular}{@{}c@{}} 113.0\\ -1.06\\ 1.00e+00\end{tabular} &     \begin{tabular}{@{}c@{}} \bf 17.0\\ \bf -3.80\\ \bf 2.26e-2\end{tabular} &     \begin{tabular}{@{}c@{}} \bf 18.0\\ \bf -3.77\\ \bf 2.53e-2\end{tabular} \\
\midrule
		\textit{ig} &      \begin{tabular}{@{}c@{}} \bf 0.0\\ \bf -4.29\\ \bf 2.84e-3\end{tabular} &      \begin{tabular}{@{}c@{}} \bf 0.0\\ \bf -4.20\\ \bf 4.21e-3\end{tabular} &      \begin{tabular}{@{}c@{}} \bf 0.0\\ \bf -4.29\\ \bf 2.84e-3\end{tabular} &      \begin{tabular}{@{}c@{}} \bf 0.0\\ \bf -4.29\\ \bf 2.84e-3\end{tabular} &      \begin{tabular}{@{}c@{}} \bf 0.0\\ \bf -4.29\\ \bf 2.84e-3\end{tabular} &      \begin{tabular}{@{}c@{}} \bf 0.0\\ \bf -4.29\\ \bf 2.84e-3\end{tabular} &      \begin{tabular}{@{}c@{}} \bf 8.0\\ \bf -3.95\\ \bf 1.20e-2\end{tabular} &                               - &  \begin{tabular}{@{}c@{}} 59.5\\ -2.18\\ 1.00e+00\end{tabular} &  \begin{tabular}{@{}c@{}} 69.5\\ -2.08\\ 1.00e+00\end{tabular} &     \begin{tabular}{@{}c@{}} \bf 20.0\\ \bf -3.71\\ \bf 3.18e-2\end{tabular} &      \begin{tabular}{@{}c@{}} \bf 0.0\\ \bf -4.29\\ \bf 2.84e-3\end{tabular} &      \begin{tabular}{@{}c@{}} \bf 0.0\\ \bf -4.29\\ \bf 2.84e-3\end{tabular} \\
\midrule
		\textit{rf} &      \begin{tabular}{@{}c@{}} \bf 0.0\\ \bf -4.29\\ \bf 2.84e-3\end{tabular} &      \begin{tabular}{@{}c@{}} \bf 0.0\\ \bf -4.29\\ \bf 2.84e-3\end{tabular} &      \begin{tabular}{@{}c@{}} \bf 0.0\\ \bf -4.29\\ \bf 2.84e-3\end{tabular} &      \begin{tabular}{@{}c@{}} \bf 0.0\\ \bf -4.29\\ \bf 2.84e-3\end{tabular} &      \begin{tabular}{@{}c@{}} \bf 0.0\\ \bf -4.29\\ \bf 2.84e-3\end{tabular} &      \begin{tabular}{@{}c@{}} \bf 0.0\\ \bf -4.29\\ \bf 2.84e-3\end{tabular} &    \begin{tabular}{@{}c@{}} 20.0\\ -3.59\\ 5.18e-2\end{tabular} &  \begin{tabular}{@{}c@{}} 59.5\\ -2.18\\ 1.00e+00\end{tabular} &                               - &  \begin{tabular}{@{}c@{}} 69.5\\ -1.03\\ 1.00e+00\end{tabular} &    \begin{tabular}{@{}c@{}} 28.0\\ -3.49\\ 7.66e-2\end{tabular} &      \begin{tabular}{@{}c@{}} \bf 0.0\\ \bf -4.29\\ \bf 2.84e-3\end{tabular} &      \begin{tabular}{@{}c@{}} \bf 0.0\\ \bf -4.29\\ \bf 2.84e-3\end{tabular} \\
\midrule
		\textit{tds} &      \begin{tabular}{@{}c@{}} \bf 0.0\\ \bf -4.29\\ \bf 2.84e-3\end{tabular} &      \begin{tabular}{@{}c@{}} \bf 0.0\\ \bf -4.29\\ \bf 2.84e-3\end{tabular} &      \begin{tabular}{@{}c@{}} \bf 0.0\\ \bf -4.29\\ \bf 2.84e-3\end{tabular} &      \begin{tabular}{@{}c@{}} \bf 0.0\\ \bf -4.29\\ \bf 2.84e-3\end{tabular} &      \begin{tabular}{@{}c@{}} \bf 0.0\\ \bf -4.29\\ \bf 2.84e-3\end{tabular} &      \begin{tabular}{@{}c@{}} \bf 0.0\\ \bf -4.29\\ \bf 2.84e-3\end{tabular} &     \begin{tabular}{@{}c@{}} \bf 21.5\\ \bf -3.67\\ \bf 3.76e-2\end{tabular} &  \begin{tabular}{@{}c@{}} 69.5\\ -2.08\\ 1.00e+00\end{tabular} &  \begin{tabular}{@{}c@{}} 69.5\\ -1.03\\ 1.00e+00\end{tabular} &                               - &    \begin{tabular}{@{}c@{}} 29.0\\ -3.46\\ 8.52e-2\end{tabular} &      \begin{tabular}{@{}c@{}} \bf 0.0\\ \bf -4.29\\ \bf 2.84e-3\end{tabular} &      \begin{tabular}{@{}c@{}} \bf 0.0\\ \bf -4.29\\ \bf 2.84e-3\end{tabular} \\
\midrule
		\textit{thd} &      \begin{tabular}{@{}c@{}} \bf 8.0\\ \bf -4.06\\ \bf 7.75e-3\end{tabular} &   \begin{tabular}{@{}c@{}} 87.0\\ -1.80\\ 1.00e+00\end{tabular} &    \begin{tabular}{@{}c@{}} 25.0\\ -3.57\\ 5.54e-2\end{tabular} &   \begin{tabular}{@{}c@{}} 89.0\\ -1.74\\ 1.00e+00\end{tabular} &    \begin{tabular}{@{}c@{}} 46.0\\ -2.97\\ 4.62e-1\end{tabular} &      \begin{tabular}{@{}c@{}} \bf 3.0\\ \bf -4.20\\ \bf 4.16e-3\end{tabular} &  \begin{tabular}{@{}c@{}} 113.0\\ -1.06\\ 1.00e+00\end{tabular} &    \begin{tabular}{@{}c@{}} \bf 20.0\\ \bf -3.71\\ \bf 3.18e-2\end{tabular} &   \begin{tabular}{@{}c@{}} 28.0\\ -3.49\\ 7.66e-2\end{tabular} &   \begin{tabular}{@{}c@{}} 29.0\\ -3.46\\ 8.52e-2\end{tabular} &                                - &    \begin{tabular}{@{}c@{}} 43.5\\ -3.04\\ 3.65e-1\end{tabular} &    \begin{tabular}{@{}c@{}} 35.0\\ -3.29\\ 1.59e-1\end{tabular} \\
\midrule
		$tf\mhyphen \chi^2$ &   \begin{tabular}{@{}c@{}} 76.0\\ -2.11\\ 1.00e+00\end{tabular} &   \begin{tabular}{@{}c@{}} 83.0\\ -1.91\\ 1.00e+00\end{tabular} &  \begin{tabular}{@{}c@{}} 136.0\\ -0.40\\ 1.00e+00\end{tabular} &   \begin{tabular}{@{}c@{}} 72.0\\ -2.23\\ 1.00e+00\end{tabular} &  \begin{tabular}{@{}c@{}} 109.0\\ -1.17\\ 1.00e+00\end{tabular} &   \begin{tabular}{@{}c@{}} 65.0\\ -2.43\\ 1.00e+00\end{tabular} &     \begin{tabular}{@{}c@{}} \bf 17.0\\ \bf -3.80\\ \bf 2.26e-2\end{tabular} &     \begin{tabular}{@{}c@{}} \bf 0.0\\ \bf -4.29\\ \bf 2.84e-3\end{tabular} &     \begin{tabular}{@{}c@{}} \bf 0.0\\ \bf -4.29\\ \bf 2.84e-3\end{tabular} &     \begin{tabular}{@{}c@{}} \bf 0.0\\ \bf -4.29\\ \bf 2.84e-3\end{tabular} &    \begin{tabular}{@{}c@{}} 43.5\\ -3.04\\ 3.65e-1\end{tabular} &                                - &  \begin{tabular}{@{}c@{}} 120.0\\ -0.86\\ 1.00e+00\end{tabular} \\
\midrule
		$\chi^2$ &  \begin{tabular}{@{}c@{}} 146.0\\ -0.11\\ 1.00e+00\end{tabular} &   \begin{tabular}{@{}c@{}} 63.0\\ -2.49\\ 1.00e+00\end{tabular} &  \begin{tabular}{@{}c@{}} 102.0\\ -1.37\\ 1.00e+00\end{tabular} &   \begin{tabular}{@{}c@{}} 64.0\\ -2.46\\ 1.00e+00\end{tabular} &   \begin{tabular}{@{}c@{}} 77.0\\ -2.09\\ 1.00e+00\end{tabular} &  \begin{tabular}{@{}c@{}} 117.0\\ -0.94\\ 1.00e+00\end{tabular} &     \begin{tabular}{@{}c@{}} \bf 18.0\\ \bf -3.77\\ \bf 2.53e-2\end{tabular} &     \begin{tabular}{@{}c@{}} \bf 0.0\\ \bf -4.29\\ \bf 2.84e-3\end{tabular} &     \begin{tabular}{@{}c@{}} \bf 0.0\\ \bf -4.29\\ \bf 2.84e-3\end{tabular} &     \begin{tabular}{@{}c@{}} \bf 0.0\\ \bf -4.29\\ \bf 2.84e-3\end{tabular} &    \begin{tabular}{@{}c@{}} 35.0\\ -3.29\\ 1.59e-1\end{tabular} &  \begin{tabular}{@{}c@{}} 120.0\\ -0.86\\ 1.00e+00\end{tabular} &                                - \\
\bottomrule
	\end{tabular}
\end{table}
		\end{landscape}
		\clearpage
	}

\afterpage{%
	\clearpage
	\thispagestyle{empty}
		\begin{landscape}

\begin{table}[htb]
	\tiny
	\caption{
		Statistical significance of micro-averaged F1 score differences for pairs of word-weighting metrics when used in our method using one-tailed sign test.
	}
	\label{table-metrics-significance2}
	\tiny
	\begin{tabular}{lcccccccccccccc}
		\toprule
		{} & \textit{tf-idf} & \textit{tf-ig} & \textit{tf-rf} & \textit{tf-tds} & \textit{tf-tf-dcf} & \textit{tf-thd} & \textit{tf-dcf} & $ig$ & $rf$ & $tds$ & $thd$ & $tf\mhyphen \chi^2$ & $\chi^2$ \\
\midrule
\midrule
		\textit{tf-idf} &                     - &   \begin{tabular}{@{}c@{}} \bf -12\\ \bf 1.86e-5\end{tabular} &  \begin{tabular}{@{}c@{}} \bf -11\\ \bf 4.65e-4\end{tabular} &   \begin{tabular}{@{}c@{}} \bf -12\\ \bf 1.86e-5\end{tabular} &   \begin{tabular}{@{}c@{}} \bf -10\\ \bf 5.60e-3\end{tabular} &   \begin{tabular}{@{}c@{}} \bf 11\\ \bf 4.65e-4\end{tabular} &   \begin{tabular}{@{}c@{}} \bf -10\\ \bf 5.60e-3\end{tabular} &   \begin{tabular}{@{}c@{}} \bf -12\\ \bf 1.86e-5\end{tabular} &   \begin{tabular}{@{}c@{}} \bf -12\\ \bf 1.86e-5\end{tabular} &  \begin{tabular}{@{}c@{}} \bf -12\\ \bf 1.86e-5\end{tabular} &   \begin{tabular}{@{}c@{}} \bf -10\\ \bf 5.60e-3\end{tabular} &  \begin{tabular}{@{}c@{}} -7\\ 1.00e+00\end{tabular} &  \begin{tabular}{@{}c@{}} -1\\ 1.00e+00\end{tabular} \\
\midrule
		\textit{tf-ig} &   \begin{tabular}{@{}c@{}} \bf 12\\ \bf 1.86e-5\end{tabular} &                      - &    \begin{tabular}{@{}c@{}} \bf 9\\ \bf 4.32e-2\end{tabular} &   \begin{tabular}{@{}c@{}} 3\\ 1.00e+00\end{tabular} &   \begin{tabular}{@{}c@{}} 3\\ 1.00e+00\end{tabular} &   \begin{tabular}{@{}c@{}} \bf 12\\ \bf 1.86e-5\end{tabular} &  \begin{tabular}{@{}c@{}} -6\\ 1.00e+00\end{tabular} &   \begin{tabular}{@{}c@{}} \bf -11\\ \bf 3.72e-5\end{tabular} &   \begin{tabular}{@{}c@{}} \bf -12\\ \bf 1.86e-5\end{tabular} &  \begin{tabular}{@{}c@{}} \bf -12\\ \bf 1.86e-5\end{tabular} &  \begin{tabular}{@{}c@{}} -4\\ 1.00e+00\end{tabular} &   \begin{tabular}{@{}c@{}} 2\\ 1.00e+00\end{tabular} &   \begin{tabular}{@{}c@{}} 4\\ 1.00e+00\end{tabular} \\
\midrule
		\textit{tf-rf} &   \begin{tabular}{@{}c@{}} \bf 11\\ \bf 4.65e-4\end{tabular} &    \begin{tabular}{@{}c@{}} \bf -9\\ \bf 4.32e-2\end{tabular} &                     - &   \begin{tabular}{@{}c@{}} \bf -10\\ \bf 5.60e-3\end{tabular} &  \begin{tabular}{@{}c@{}} -4\\ 1.00e+00\end{tabular} &   \begin{tabular}{@{}c@{}} \bf 12\\ \bf 1.86e-5\end{tabular} &   \begin{tabular}{@{}c@{}} -8\\ 2.41e-1\end{tabular} &   \begin{tabular}{@{}c@{}} \bf -12\\ \bf 1.86e-5\end{tabular} &   \begin{tabular}{@{}c@{}} \bf -12\\ \bf 1.86e-5\end{tabular} &  \begin{tabular}{@{}c@{}} \bf -12\\ \bf 1.86e-5\end{tabular} &  \begin{tabular}{@{}c@{}} -7\\ 1.00e+00\end{tabular} &  \begin{tabular}{@{}c@{}} -1\\ 1.00e+00\end{tabular} &   \begin{tabular}{@{}c@{}} 0\\ 1.00e+00\end{tabular} \\
\midrule
		\textit{tf-tds} &   \begin{tabular}{@{}c@{}} \bf 12\\ \bf 1.86e-5\end{tabular} &  \begin{tabular}{@{}c@{}} -3\\ 1.00e+00\end{tabular} &   \begin{tabular}{@{}c@{}} \bf 10\\ \bf 5.60e-3\end{tabular} &                      - &   \begin{tabular}{@{}c@{}} 5\\ 1.00e+00\end{tabular} &   \begin{tabular}{@{}c@{}} \bf 12\\ \bf 1.86e-5\end{tabular} &  \begin{tabular}{@{}c@{}} -6\\ 1.00e+00\end{tabular} &   \begin{tabular}{@{}c@{}} \bf -12\\ \bf 1.86e-5\end{tabular} &   \begin{tabular}{@{}c@{}} \bf -12\\ \bf 1.86e-5\end{tabular} &  \begin{tabular}{@{}c@{}} \bf -12\\ \bf 1.86e-5\end{tabular} &  \begin{tabular}{@{}c@{}} -4\\ 1.00e+00\end{tabular} &   \begin{tabular}{@{}c@{}} 3\\ 1.00e+00\end{tabular} &   \begin{tabular}{@{}c@{}} 4\\ 1.00e+00\end{tabular} \\
\midrule
		\textit{tf-tf-dcf} &   \begin{tabular}{@{}c@{}} \bf 10\\ \bf 5.60e-3\end{tabular} &  \begin{tabular}{@{}c@{}} -3\\ 1.00e+00\end{tabular} &  \begin{tabular}{@{}c@{}} 4\\ 1.00e+00\end{tabular} &  \begin{tabular}{@{}c@{}} -5\\ 1.00e+00\end{tabular} &                      - &   \begin{tabular}{@{}c@{}} \bf 11\\ \bf 4.65e-4\end{tabular} &  \begin{tabular}{@{}c@{}} -6\\ 1.00e+00\end{tabular} &   \begin{tabular}{@{}c@{}} \bf -12\\ \bf 1.86e-5\end{tabular} &   \begin{tabular}{@{}c@{}} \bf -12\\ \bf 1.86e-5\end{tabular} &  \begin{tabular}{@{}c@{}} \bf -12\\ \bf 1.86e-5\end{tabular} &  \begin{tabular}{@{}c@{}} -4\\ 1.00e+00\end{tabular} &   \begin{tabular}{@{}c@{}} 3\\ 1.00e+00\end{tabular} &   \begin{tabular}{@{}c@{}} 2\\ 1.00e+00\end{tabular} \\
\midrule
		\textit{tf-thd} &  \begin{tabular}{@{}c@{}} \bf -11\\ \bf 4.65e-4\end{tabular} &   \begin{tabular}{@{}c@{}} \bf -12\\ \bf 1.86e-5\end{tabular} &  \begin{tabular}{@{}c@{}} \bf -12\\ \bf 1.86e-5\end{tabular} &   \begin{tabular}{@{}c@{}} \bf -12\\ \bf 1.86e-5\end{tabular} &   \begin{tabular}{@{}c@{}} \bf -11\\ \bf 4.65e-4\end{tabular} &                     - &   \begin{tabular}{@{}c@{}} \bf -11\\ \bf 4.65e-4\end{tabular} &   \begin{tabular}{@{}c@{}} \bf -12\\ \bf 1.86e-5\end{tabular} &   \begin{tabular}{@{}c@{}} \bf -12\\ \bf 1.86e-5\end{tabular} &  \begin{tabular}{@{}c@{}} \bf -12\\ \bf 1.86e-5\end{tabular} &   \begin{tabular}{@{}c@{}} \bf -11\\ \bf 4.65e-4\end{tabular} &  \begin{tabular}{@{}c@{}} -7\\ 1.00e+00\end{tabular} &  \begin{tabular}{@{}c@{}} -1\\ 1.00e+00\end{tabular} \\
\midrule
		\textit{tf-dcf} &   \begin{tabular}{@{}c@{}} \bf 10\\ \bf 5.60e-3\end{tabular} &   \begin{tabular}{@{}c@{}} 6\\ 1.00e+00\end{tabular} &   \begin{tabular}{@{}c@{}} 8\\ 2.41e-1\end{tabular} &   \begin{tabular}{@{}c@{}} 6\\ 1.00e+00\end{tabular} &   \begin{tabular}{@{}c@{}} 6\\ 1.00e+00\end{tabular} &   \begin{tabular}{@{}c@{}} \bf 11\\ \bf 4.65e-4\end{tabular} &                      - &    \begin{tabular}{@{}c@{}} \bf -9\\ \bf 1.03e-2\end{tabular} &   \begin{tabular}{@{}c@{}} -7\\ 4.06e-1\end{tabular} &   \begin{tabular}{@{}c@{}} \bf -9\\ \bf 4.32e-2\end{tabular} &   \begin{tabular}{@{}c@{}} 2\\ 1.00e+00\end{tabular} &     \begin{tabular}{@{}c@{}} \bf 9\\ \bf 4.32e-2\end{tabular} &    \begin{tabular}{@{}c@{}} \bf 10\\ \bf 5.60e-3\end{tabular} \\
\midrule
		\textit{ig} &   \begin{tabular}{@{}c@{}} \bf 12\\ \bf 1.86e-5\end{tabular} &    \begin{tabular}{@{}c@{}} \bf 11\\ \bf 3.72e-5\end{tabular} &   \begin{tabular}{@{}c@{}} \bf 12\\ \bf 1.86e-5\end{tabular} &    \begin{tabular}{@{}c@{}} \bf 12\\ \bf 1.86e-5\end{tabular} &    \begin{tabular}{@{}c@{}} \bf 12\\ \bf 1.86e-5\end{tabular} &   \begin{tabular}{@{}c@{}} \bf 12\\ \bf 1.86e-5\end{tabular} &     \begin{tabular}{@{}c@{}} \bf 9\\ \bf 1.03e-2\end{tabular} &                      - &   \begin{tabular}{@{}c@{}} 6\\ 1.00e+00\end{tabular} &  \begin{tabular}{@{}c@{}} 5\\ 1.00e+00\end{tabular} &    \begin{tabular}{@{}c@{}} \bf 10\\ \bf 5.60e-3\end{tabular} &    \begin{tabular}{@{}c@{}} \bf 12\\ \bf 1.86e-5\end{tabular} &    \begin{tabular}{@{}c@{}} \bf 12\\ \bf 1.86e-5\end{tabular} \\
\midrule
		\textit{rf} &   \begin{tabular}{@{}c@{}} \bf 12\\ \bf 1.86e-5\end{tabular} &    \begin{tabular}{@{}c@{}} \bf 12\\ \bf 1.86e-5\end{tabular} &   \begin{tabular}{@{}c@{}} \bf 12\\ \bf 1.86e-5\end{tabular} &    \begin{tabular}{@{}c@{}} \bf 12\\ \bf 1.86e-5\end{tabular} &    \begin{tabular}{@{}c@{}} \bf 12\\ \bf 1.86e-5\end{tabular} &   \begin{tabular}{@{}c@{}} \bf 12\\ \bf 1.86e-5\end{tabular} &    \begin{tabular}{@{}c@{}} 7\\ 4.06e-1\end{tabular} &  \begin{tabular}{@{}c@{}} -6\\ 1.00e+00\end{tabular} &                      - &  \begin{tabular}{@{}c@{}} 1\\ 1.00e+00\end{tabular} &     \begin{tabular}{@{}c@{}} \bf 9\\ \bf 4.32e-2\end{tabular} &    \begin{tabular}{@{}c@{}} \bf 12\\ \bf 1.86e-5\end{tabular} &    \begin{tabular}{@{}c@{}} \bf 12\\ \bf 1.86e-5\end{tabular} \\
\midrule
		\textit{tds} &   \begin{tabular}{@{}c@{}} \bf 12\\ \bf 1.86e-5\end{tabular} &    \begin{tabular}{@{}c@{}} \bf 12\\ \bf 1.86e-5\end{tabular} &   \begin{tabular}{@{}c@{}} \bf 12\\ \bf 1.86e-5\end{tabular} &    \begin{tabular}{@{}c@{}} \bf 12\\ \bf 1.86e-5\end{tabular} &    \begin{tabular}{@{}c@{}} \bf 12\\ \bf 1.86e-5\end{tabular} &   \begin{tabular}{@{}c@{}} \bf 12\\ \bf 1.86e-5\end{tabular} &     \begin{tabular}{@{}c@{}} \bf 9\\ \bf 4.32e-2\end{tabular} &  \begin{tabular}{@{}c@{}} -5\\ 1.00e+00\end{tabular} &  \begin{tabular}{@{}c@{}} -1\\ 1.00e+00\end{tabular} &                     - &     \begin{tabular}{@{}c@{}} \bf 9\\ \bf 4.32e-2\end{tabular} &    \begin{tabular}{@{}c@{}} \bf 12\\ \bf 1.86e-5\end{tabular} &    \begin{tabular}{@{}c@{}} \bf 12\\ \bf 1.86e-5\end{tabular} \\
\midrule
		\textit{thd} &   \begin{tabular}{@{}c@{}} \bf 10\\ \bf 5.60e-3\end{tabular} &   \begin{tabular}{@{}c@{}} 4\\ 1.00e+00\end{tabular} &  \begin{tabular}{@{}c@{}} 7\\ 1.00e+00\end{tabular} &   \begin{tabular}{@{}c@{}} 4\\ 1.00e+00\end{tabular} &   \begin{tabular}{@{}c@{}} 4\\ 1.00e+00\end{tabular} &   \begin{tabular}{@{}c@{}} \bf 11\\ \bf 4.65e-4\end{tabular} &  \begin{tabular}{@{}c@{}} -2\\ 1.00e+00\end{tabular} &   \begin{tabular}{@{}c@{}} \bf -10\\ \bf 5.60e-3\end{tabular} &    \begin{tabular}{@{}c@{}} \bf -9\\ \bf 4.32e-2\end{tabular} &   \begin{tabular}{@{}c@{}} \bf -9\\ \bf 4.32e-2\end{tabular} &                      - &   \begin{tabular}{@{}c@{}} 7\\ 1.00e+00\end{tabular} &    \begin{tabular}{@{}c@{}} 8\\ 2.41e-1\end{tabular} \\
\midrule
		$tf\mhyphen \chi^2$ &  \begin{tabular}{@{}c@{}} 7\\ 1.00e+00\end{tabular} &  \begin{tabular}{@{}c@{}} -2\\ 1.00e+00\end{tabular} &  \begin{tabular}{@{}c@{}} 1\\ 1.00e+00\end{tabular} &  \begin{tabular}{@{}c@{}} -3\\ 1.00e+00\end{tabular} &  \begin{tabular}{@{}c@{}} -3\\ 1.00e+00\end{tabular} &  \begin{tabular}{@{}c@{}} 7\\ 1.00e+00\end{tabular} &    \begin{tabular}{@{}c@{}} \bf -9\\ \bf 4.32e-2\end{tabular} &   \begin{tabular}{@{}c@{}} \bf -12\\ \bf 1.86e-5\end{tabular} &   \begin{tabular}{@{}c@{}} \bf -12\\ \bf 1.86e-5\end{tabular} &  \begin{tabular}{@{}c@{}} \bf -12\\ \bf 1.86e-5\end{tabular} &  \begin{tabular}{@{}c@{}} -7\\ 1.00e+00\end{tabular} &                      - &   \begin{tabular}{@{}c@{}} 0\\ 1.00e+00\end{tabular} \\
\midrule
		$\chi^2$ &  \begin{tabular}{@{}c@{}} 1\\ 1.00e+00\end{tabular} &  \begin{tabular}{@{}c@{}} -4\\ 1.00e+00\end{tabular} &  \begin{tabular}{@{}c@{}} 0\\ 1.00e+00\end{tabular} &  \begin{tabular}{@{}c@{}} -4\\ 1.00e+00\end{tabular} &  \begin{tabular}{@{}c@{}} -2\\ 1.00e+00\end{tabular} &  \begin{tabular}{@{}c@{}} 1\\ 1.00e+00\end{tabular} &   \begin{tabular}{@{}c@{}} \bf -10\\ \bf 5.60e-3\end{tabular} &   \begin{tabular}{@{}c@{}} \bf -12\\ \bf 1.86e-5\end{tabular} &   \begin{tabular}{@{}c@{}} \bf -12\\ \bf 1.86e-5\end{tabular} &  \begin{tabular}{@{}c@{}} \bf -12\\ \bf 1.86e-5\end{tabular} &   \begin{tabular}{@{}c@{}} -8\\ 2.41e-1\end{tabular} &   \begin{tabular}{@{}c@{}} 0\\ 1.00e+00\end{tabular} &                      - \\
\bottomrule
	\end{tabular}
\end{table}
		\end{landscape}
		\clearpage
	}

	\clearpage

	\section{Original texts used to extract example keywords}
	\label{appendix-sample-documents}

	\tiny
	\subsection{First document in category \textit{comp.graphics} of 20newsgroups dataset}
	From: ldo@waikato.ac.nz (Lawrence D'Oliveiro, Waikato University)

	Subject: QuickTime performance (was Re: Rumours about 3DO ???)

	Organization: University of Waikato, Hamilton, New Zealand

	Lines: 67

	OK, with all the discussion about observed playback speeds with QuickTime,
	the effects of scaling and so on, I thought I'd do some more tests.

	First of all, I felt that my original speed test was perhaps less than
	realistic. The movie I had been using only had 18 frames in it (it was a
	version of the very first movie I created with the Compact Video compressor).
	I decided something a little longer would give closer to real-world results
	(for better or for worse).

	I pulled out a copy of "2001: A Space Odyssey" that I had recorded off TV
	a while back. About fifteen minutes into the movie, there's a sequence where
	the Earth shuttle is approaching the space station. Specifically, I digitized
	a portion of about 30 seconds' duration, zooming in on the rotating space
	station. I figured this would give a reasonable amount of movement between
	frames. To increase the differences between frames, I digitized it at only
	5 frames per second, to give a total of 171 frames.

	I captured the raw footage at a resolution of 384*288 pixels with the Spigot
	card in my Centris 650 (quarter-size resolution from a PAL source). I then
	imported it into Premiere and put it through the Compact Video compressor,
	keeping the 5 fps frame rate. I created two versions of the movie: one scaled
	to 320*240 resolution, the other at 160*120 resolution. I used the default
	"2.00" quality setting in Premiere 2.0.1, and specified a key frame every ten
	frames.

	I then ran the 320*240 movie through the same "Raw Speed Test" program I used
	for the results I'd been reporting earlier.

	Result: a playback rate of over 45 frames per second.

	That's right, I was getting a much higher result than with that first short
	test movie. Just for fun, I copied the 320*240 movie to my external hard
	disk (a Quantum LP105S), and ran it from there. This time the playback rate
	was only about 35 frames per second. Obviously the 230MB internal hard disk
	(also a Quantum) is a significant contributor to the speed of playback.

	I modified my speed test program to allow the specification of optional
	scaling factors, and tried playing back the 160*120 movie scaled to 320*240
	size. This time the playback speed was over 60 fps. Clearly, the poster who
	observed poor performance on scaled playback was seeing QuickTime 1.0 in
	action, not 1.5. I'd try my tests with QuickTime 1.0, but I don't think it's
	entirely compatible with my Centris and System 7.1...

	Unscaled, the playback rate for the 160*120 movie was over 100 fps.

	The other thing I tried was saving versions of the 320*240 movie with
	"preferred" playback rates greater than 1.0, and seeing how well they played
	from within MoviePlayer (ie with QuickTime's normal synchronized playback).
	A preferred rate of 9.0 (=> 45 fps) didn't work too well: the playback was
	very jerky. Compare this with the raw speed test, which achieved 45 fps with
	ease. I can't believe that QuickTime's synchronization code would add this
	much overhead: I think the slowdown was coming from the Mac system's task
	switching.

	A preferred rate of 7.0 (=> 35 fps) seemed to work fine: I couldn't see
	any evidence of stutter. At 8.0 (=> 40 fps) I *think* I could see slight
	stutter, but with four key frames every second, it was hard to tell.

	I guess I could try recreating the movies with a longer interval between the
	key frames, to make the stutter more noticeable. Of course, this will also
	improve the compression slightly, which should speed up the playback performance
	even more...

	Lawrence D'Oliveiro                       fone: +64-7-856-2889
	Computer Services Dept                     fax: +64-7-838-4066
	University of Waikato            electric mail: ldo@waikato.ac.nz
	Hamilton, New Zealand    37\^ 47' 26" S, 175\^ 19' 7" E, GMT+12:00

	\subsection{First document in category \textit{ship} of Reuters-21578 dataset}
	AUSTRALIAN FOREIGN SHIP BAN ENDS BUT NSW PORTS HIT
	SYDNEY, April 8 - Tug crews in New South Wales (NSW),
	Victoria and Western Australia yesterday lifted their ban on
	foreign-flag ships carrying containers but NSW ports are still
	being disrupted by a separate dispute, shipping sources said.
	The ban, imposed a week ago over a pay claim, had prevented
	the movement in or out of port of nearly 20 vessels, they said.
	The pay dispute went before a hearing of the Arbitration
	Commission today.
	Meanwhile, disruption began today to cargo handling in the
	ports of Sydney, Newcastle and Port Kembla, they said.
	The industrial action at the NSW ports is part of the week
	of action called by the NSW Trades and Labour Council to
	protest changes to the state's workers' compensation laws.
	The shipping sources said the various port unions appear to
	be taking it in turn to work for a short time at the start of
	each shift and then to walk off.
	Cargo handling in the ports has been disrupted, with
	container movements most affected, but has not stopped
	altogether, they said.
	They said they could not say how long the disruption will
	go on and what effect it will have on shipping movements.
	REUTER




	%
	%
	%
	\end{document}